\documentclass[letterpaper,11pt,twoside]{article}

\usepackage{jair}
\usepackage{theapa}

\usepackage{bcprules}
\usepackage{ASL-ASC-JAIR}
\usepackage{graphicx}
\usepackage{alltt}
\usepackage{latexsym}
\usepackage{url}

\newenvironment{paragraph*}{\vspace*{1ex}\noindent\textbf}{}

\newtheorem{defn}{Definition}
\newtheorem{exmp}{Example}
\newtheorem{prop}{Proposition}
\newenvironment{pf}{\medskip\noindent\textbf{Proof.\quad}}{\hspace*{\fill}\qed\\\medskip}
\newcommand{\qed}{\unskip\kern 10pt{\unitlength1pt\linethickness{.4pt}\framebox(6,6){}}}

\jairheading{29}{2007}{221--267}{12/06}{06/07}
\ShortHeadings{Speech-Act Based Communication in Agent
  Programming}{Vieira, Moreira, Wooldridge, \& Bordini}
\firstpageno{221}

\begin{document}
\title{On the Formal Semantics of\\
       Speech-Act Based Communication in an\\
       Agent-Oriented Programming Language
}

\author{\name Renata Vieira \email renatav@unisinos.br\\
        \addr Universidade do Vale do Rio dos Sinos \\ S\~ao Leopoldo, RS, 93022-000, Brazil
        \AND
       \name {\'A}lvaro  Moreira \email Alvaro.Moreira@inf.ufrgs.br\\
       \addr Universidade Federal do Rio Grande do Sul \\ Porto Alegre, RS, 91501-970, Brazil
       \AND
       \name Michael Wooldridge \email mjw@csc.liv.ac.uk\\
       \addr University of Liverpool \\ Liverpool L69 3BX, United
       Kingdom
       \AND
       \name  Rafael H. Bordini \email R.Bordini@durham.ac.uk\\
       \addr University of Durham \\ Durham DH1 3LE, United Kingdom}


\maketitle

\begin{abstract}\noindent
  Research on agent communication languages has typically taken the
  speech acts paradigm as its starting point.  Despite their manifest
  attractions, speech-act models of communication have several serious
  disadvantages as a foundation for communication in artificial agent
  systems. In particular, it has proved to be extremely difficult to
  give a satisfactory semantics to speech-act based agent
  communication languages.  In part, the problem is that speech-act
  semantics typically make reference to the ``mental states'' of
  agents (their beliefs, desires, and intentions), and there is in
  general no way to attribute such attitudes to arbitrary
  computational agents.  In addition, agent programming languages have
  only had their semantics formalised for abstract, stand-alone
  versions, neglecting aspects such as communication primitives. With
  respect to communication, implemented agent programming languages
  have tended to be rather \emph{ad hoc}. This paper addresses both of
  these problems, by giving semantics to speech-act based messages
  received by an \as\ agent. \as\ is a logic-based agent programming
  language which incorporates the main features of the PRS model of
  reactive planning systems.  The paper builds upon a structural
  operational semantics to \as\ that we developed in previous work.
  The main contributions of this paper are as follows: an extension of
  our earlier work on the theoretical foundations of \as\
  interpreters; a computationally grounded semantics for (the core)
  performatives used in speech-act based agent communication
  languages; and a well-defined extension of \as\ that supports agent
  communication.
\end{abstract}

\section{Introduction}
\label{secI}

First introduced in 1987, the reactive planning model of Georgeff and
Lansky's PRS system has subsequently proved to be one of the most
influential and long-lived approaches to programming multi-agent
systems \cite{GeorRRP}.  The \as\ programming language, introduced by
Rao \citeyear{RaoASLBDIASOLCL}, represents an attempt to distill the
key features of the PRS approach into a simple, abstract, logic-based
language. \as\ is particularly interesting, in comparison to other
agent-oriented languages, in that it retains the most important
aspects of the BDI-based reactive planning systems on which it was
based, and at the same time it has robust working interpreters
\cite{BordiniJGFAOP,BordiniJASON,BordiniASXLeisbadtts}, its formal
semantics and relation to BDI logics \cite{RaoDPBDIL,WoolRRA} have
been thoroughly studied
\cite{BordiniPBPAOPL-ATPA,MoreiraEOSBAOPLISABC-LNAI,MoreiraOSBAOPL},
and there is ongoing work on the use of model-checking techniques for
verification of \as\ multi-agent systems
\cite{BordiniMCRA,BordiniMCMAP,BordiniMCAS}.

In the original formulation of \as \cite{RaoASLBDIASOLCL}, the main
emphasis was on the \emph{internal} control structures and
decision-making cycle of an agent: the issue of \emph{communication}
between agents was not addressed. Accordingly, most attempts to give a
formal semantics to the language have focused on these internal
aspects \cite{MoreiraOSBAOPL}.  Although several extensions to \as\
have been proposed in an attempt to make it a practically more useful
language \cite{BordiniJGFAOP,BordiniASXLeisbadtts}, comparatively
little research has addressed the issue of a principled mechanism to
support communication in \as, which is clearly essential for
engineering \emph{multi}-agent systems.

Most agent communication languages have taken \emph{speech-act
  theory}~\cite{AustinHDTW,SearleSAEPL} as their starting point.  As
is suggested by its name, speech-act theory is predicated on the view
that utterances are \emph{actions}, performed by rational agents in
the furtherance of their personal desires and intentions.  Thus,
according to speech-act theory, utterances may be considered as
actions performed by an agent, typically with the intention of
changing the mental state of the hearer(s) of the utterance.
Speech-act theory thus seems particularly appropriate as a foundation
for communication among intentional agents. Through communication, an
agent can share its internal state (beliefs, desires, intentions) with
other agents, and can attempt to influence the mental states of other
agents.

Although an initial speech-act based communication model for \as\ 
agents was previously introduced \cite{BordiniMCAS}, no formal
semantics of that model was given in that paper. A preliminary formal
account for communication of \as agents was first given by
\citeauthor{MoreiraEOSBAOPLISABC-LNAI}
\citeyear{MoreiraEOSBAOPLISABC-LNAI}. The main contribution of the
present paper is to thoroughly extend the operational semantics of \as
accounting for speech-act style communication. Our semantics precisely
defines how to implement the processing of messages \emph{received} by
an \as\ agent; that is, how the computational representations of
mental states are changed when a message is received. Note that in
implementations of the BDI architecture, the concepts of \emph{plan}
and \emph{plan library} is used to simplify aspects of deliberation
and means-ends reasoning. Therefore, an \as\ agent \emph{sends} a
message whenever there is a communicative action in the body of an
intended plan that is being executed; such plans are typically written
by an agent programmer.

As pointed out by Singh \citeyear{SinghACLRP}, well-known approaches
to agent communication focus largely on the sender's perspective,
ignoring how a message should be processed and understood. This is the
main aspect of agent communication that we consider in this paper. In
extending the operational semantics of \as\ to account for inter-agent
communication, we also touch upon another long-standing problem in the
area of multi-agent systems: the semantics of communication languages
based on speech acts. The difficulty here is that, taking their
inspiration from attempts to develop a semantics of human speech acts,
most semantics for agent communication languages have defined the
meaning of messages between agents with respect to the mental states
of communication participants. While this arguably has the advantage
of remaining neutral on the actual internal structure of agents, a
number of authors have observed that this makes it impossible in
general to determine whether or not some program that claims to be
implementing the semantics really \emph{is} implementing it
\cite{Wool98,SinghACLRP}. The problem is that if the semantics makes
reference to an agent believing (or intending a state satisfying) a
certain proposition, there is no way to ensure that any software using
that communication language complies with the underlying semantics of
belief (or intention, or mental attitudes in general).

This is related to the fact that previous approaches attempt to give a
programming language independent semantics of agent communication. Our
semantics, while developed for one specific language, have the
advantage of not relying on mechanisms --- such as abstractly defined
mental states --- that cannot be verified for real programs. We note
that, to the best of our knowledge, our work represents the first
semantics given for a speech-act style, ``knowledge level''
communication language that is used in a real system.

Since a precise notion of Belief-Desire-Intention has been given
previously for \as\ agents \cite{BordiniPBPAOPL-ATPA}, we can provide
such a computationally grounded \cite{WoolCGTA} semantics of
speech-act based communication for this language, making it possible
to determine how an \as\ agent interprets a particular message when it
is received. Note, however, that whether and how an agent acts upon
received communication depends on its plan library and its other
circumstances at the time the message is processed. Also, although our
approach is tied to a particular language, it can be usefully employed
as a reference model for developing communication semantics and
implementing communication in other agent programming languages.

The remainder of this paper is organised as follows.
Section~\ref{secBSABAC} provides the general background on PRS-style
BDI architectures and speech-act based agent communication.
Section~\ref{secSSA} presents \as syntax and semantics --- a much
revised version of the syntax and semantics of \as\ presented by
Moreira and Bordini \citeyear{MoreiraOSBAOPL,BordiniPBPAOPL-ATPA}.
Section~\ref{secSCAA} presents the speech-act based communication
model for \as\ agents, an extension of the preliminary formal account
given by \citeauthor{MoreiraEOSBAOPLISABC-LNAI}
\citeyear{MoreiraEOSBAOPLISABC-LNAI}. Section~\ref{secERARC}
illustrates the semantics with an example of the semantic rules
applied in a typical reasoning cycle. In Section~\ref{secDEFC}, we
show how programmers can use our basic communication constructs to
develop some of the more elaborate forms of communication required by
some multi-agent applications (for example, ensuring that a belief is
shared between two agents and keeping track of the progress in the
achievement of a delegated goal), and in Section~\ref{secPCPAA} we
give a simple example of the use of our framework for proving
properties of communicating agents. Section~\ref{secApp} presents a
discussion on applications and further developments for the language
presented in this paper.  Conclusions and planned future work are
given in the final section.

\section{Background}
\label{secBSABAC}

The ability to \emph{plan} seems to be one of the key components of
rational action in humans. Planning is the ability to take a goal, and
from this goal generate a ``recipe'' (i.e., plan) for action such
that, if this recipe is followed (under favourable conditions), the
goal will be achieved. Accordingly, a great deal of research in
artificial intelligence has addressed the issue of \emph{automatic
  planning}: the synthesis of plans by agents from first
principles \cite{Allen:Readings}.  Unfortunately, planning is, like so
many other problems in artificial intelligence, prohibitively
expensive in computational terms. While great strides have been made
in developing efficient automatic planning
systems \cite{ghallab:2004a}, the inherent complexity of the process
inevitably casts some doubt on whether it will be possible to use
plan-synthesis algorithms to develop plans at run-time in systems that
must operate under tight real-time constraints. Many researchers have
instead considered approaches that make use of \emph{pre-compiled}
plans, i.e., plans developed off-line, at design time. The
\emph{Procedural Reasoning System (PRS)} of Georgeff and Lansky is a
common ancestor of many such approaches \cite{GeorRRP}.

\subsection{The PRS and AgentSpeak}

\begin{figure}[thb]
\begin{center}
\fbox{\includegraphics[width=.98\linewidth]{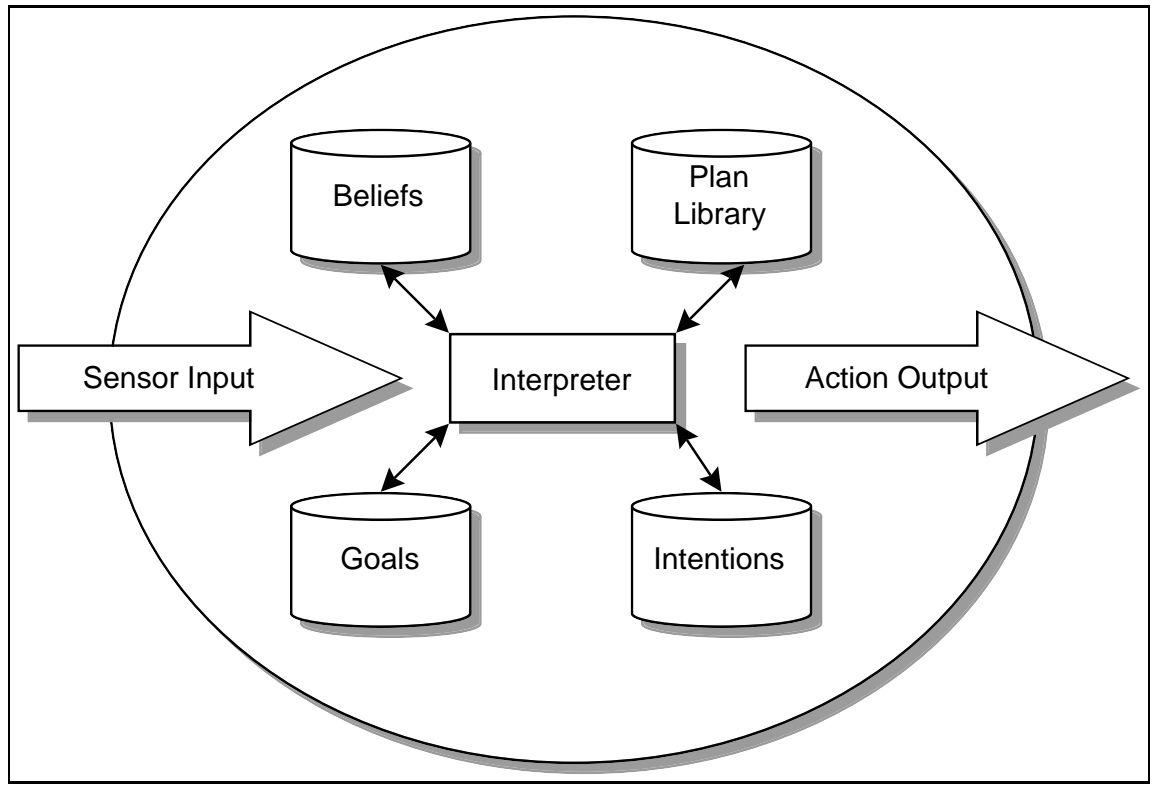}}
\caption{The PRS architecture.}
\label{fig:prs}
\end{center}
\end{figure}

On one level, the PRS can be understood simply as an architecture for
executing pre-compiled plans. However, the control structures in the
architecture incorporate a number of features which together provide a
sophisticated environment for run-time practical reasoning. First,
plans may be invoked by their \emph{effect}, rather than simply by
name (as is the case in conventional programming languages). Second,
plans are associated with a \emph{context}, which must match the
agent's current situation in order for the plan to be considered a
viable option. These two features mean that an agent may have multiple
potential plans for the same end, and can dynamically select between
these at run-time, depending on current circumstances. In addition,
plans are associated with \emph{triggering events}, the idea being
that a plan is made ``active'' by the occurrence of such an event,
which may be external or internal to the agent. External events are
changes in the environment as perceived by the agent; an example of an
internal event might be the creation of a new sub-goal, or the failure
of a plan to achieve its desired effect. Thus, overall, plans may be
invoked in a goal-driven manner (to satisfy a sub-goal that has been
created) or in an event-driven manner. The PRS architecture is
illustrated in Figure~\ref{fig:prs}. The \as\ language, introduced by
Rao \citeyear{RaoASLBDIASOLCL}, represents an attempt to distill the
``essential'' features of the PRS into a simple, unified programming
language\footnote{The name of the language originally introduced by
  Rao \citeyear{RaoASLBDIASOLCL} was \ASL. In this paper, we adopt the
  simpler form \as\ instead, and we use it to refer both to the
  original language and the variants that appeared in the
  literature.}; we provide a detailed introduction to \as\ below,
after we discuss speech-act theory and agent communication.

\subsection{Speech Acts}

The PRS model, and the \as\ language in turn, are primarily concerned
with the internal structure of decision making, and in particular the
interplay between the creation of \mbox{(sub-)goals} and the execution
of plans to achieve these \mbox{(sub-)goals}. The twin issues of
\emph{communication} and \emph{multi-agent interaction} are not
addressed within the basic architecture. This raises the question of
how such issues might be dealt with within the architecture. While BDI
theory is based on the philosophical literature on practical reasoning
\cite{BratmanIPPR}, agent communication in multi-agent systems is
typically based on the speech-act theory, in particular the work of
Austin \citeyear{AustinHDTW} and Searle \citeyear{SearleSAEPL}.

Speech-act theory starts from the principle that language is action: a
rational agent makes an utterance in an attempt to change the state of
the world, in the same way that an agent performs ``physical'' actions
to change the state of the world. What distinguishes speech acts from
other (``non-speech'') actions is that the domain of a speech act ---
the part of the world that the agent wishes to modify through the
performance of the act --- is mostly the mental state(s) of the
hearer(s) of the utterance.

Speech acts are generally classified according to their
\emph{illocutionary force} --- the ``type'' of the utterance. In
natural language, illocutionary forces are associated to utterances
(or locutionary acts).  The utterance ``the door is open'', for
example, is generally an ``inform'' or ``tell'' type of action. The
\emph{perlocutionary force} represents what the speaker of the
utterance is attempting to achieve by performing the act. In making a
statement such as ``open the door'', the perlocutionary force will
generally be the state of affairs that the speaker hopes to bring
about by making the utterance; of course, the \emph{actual} effect of
an utterance will be beyond the control of the speaker.  Whether I
choose to believe you when you inform me that the door is open depends
upon how I am disposed towards you. In natural language, the
illocutionary force and perlocutionary force will be implicit within
the speech act and its context. When the theory is adapted to agent
communication, however, the illocutionary forces are made explicit to
facilitate processing the communication act. The various types of
speech acts are generally referred to as ``performatives'' in the
context of agent communication.

Other pragmatic factors related to communication such as social roles
and conventions have been discussed in the literature
\cite{LeviEISAMD,BallmerSAC,SinghMSTFIKHC}. Illocutionary forces may
require the existence of certain relationships between speaker and
hearer for them to be felicitous. A \emph{command}, for instance,
requires a subordination relation between the individuals involved in
the communication, whereas such subordination is not required in a
\emph{request}.

Apart from illocutionary forces and social roles, other
classifications of the relations among speech acts have been proposed
\cite{LeviEISAMD}; for example, a reply follows a question, and
threatening is stronger than warning. Such categories place messages
in the larger context of a multi-agent dialogue. In multi-agent
systems, communicative interactions can be seen as communication
protocols, which in turn are normally related to a specific
coordination/cooperation mechanism. The Contract Net
\cite{SmithCNPHLCCDPS}, for example, is a protocol for task
allocation, which is defined in terms of a number of constituent
performatives (such as announcing and bidding).

\subsection{Agent Communication Languages: KQML \& FIPA}

The Knowledge Query and Manipulation Language (KQML), developed in the
context of the ``Knowledge Sharing Effort'' project \cite{GeneSA}, was
the first attempt to define a practical agent communication language
that included high level (speech-act based) communication as
considered in the distributed artificial intelligence literature. KQML
is essentially a knowledge-level messaging language
\cite{LabrouSAKQML,MayfieldEKQMLACL}. KQML defines a number of
performatives, which make explicit an agent's intentions in sending a
message. For example, the KQML performative \texttt{tell} is used with
the intention of changing the receiver's \emph{beliefs}, whereas
\texttt{achieve} is used with the intention of changing the receiver's
\emph{goals}. Thus the performative label of a KQML message explicitly
identifies the intent of the message sender.

The FIPA standard for agent
communication\footnote{http://www.fipa.org/specs/fipa00037/SC00037J.html}
was released in 2002. This standard is closely based on KQML, being
almost identical conceptually and syntactically, while differing in
the performative set and certain details of the semantic framework
\cite{LabrouCLACL}. These differences are not important for the
purposes of this paper; when we refer to traditional approaches to
semantics of speech-act based inter-agent communication, the reference
applies to both equally. However, for historical reasons, we refer
mainly to KQML and the richer literature that can be found on its
semantics.

\subsection{The Semantics of Agent Communication Languages}

Perhaps the first serious attempt to define the semantics of KQML was
made by Labrou and Finin \citeyear{LabrouSAKQML}.  Their work built on
the pioneering work of Cohen and Perrault on an action-theoretic
semantics of natural language speech acts \cite{cohen:79a}. The key
insight in Cohen and Perrault's work was that, if we take seriously
the idea of utterances as action, then we should be able to apply a
formalism for reasoning about action to reasoning about
utterances. They used a STRIPS-style pre- and post-condition formalism
to define the semantics of ``inform'' and ``request'' speech acts
(perhaps the canonical examples of speech acts), where these pre- and
post-conditions were framed in terms of the beliefs, desires, and
abilities of conversation participants. When applied by Labrou and
Finin to the KQML language \citeyear{LabrouSAKQML}, the pre- and
post-conditions defined the mental states of the sender and receiver
of a KQML message before and after sending such message. For the
description of mental states, most of the work in the area is based on
Cohen and Levesque's theory of intention
\citeyear{CohenICC,CohenRIBC}. Agent states are described through
mental attitudes such as belief ($bel$), knowledge ($know$), desire
($want$), and intention ($intend$). These mental attitudes normally
have propositions (i.e., symbolic representations of states of the
world) as arguments. Figures~\ref{fig:LST} and~\ref{fig:LSA} give
semantics for the KQML performatives $tell(S,R,X)$ ($S$ tells $R$ that
$S$ believes that $X$ is true), and $ask\_if(S,R,X)$ ($S$ asks $R$ if
$R$ believes that $X$ is true), in the style introduced by Labrou and
Finin \citeyear{LabrouSAKQML}.
\begin{figure}[!ht]
\centering \fbox{\parbox{.98\linewidth}{ \vspace{-.75\baselineskip}
\small\ttfamily
\begin{itemize}
\item
Pre-conditions on the states of $S$ and $R$:

\begin{itemize}
\item $Pre(S)$: $bel(S,X) \wedge
know(S,want(R,know(R,bel(S,X))))$
 \item $Pre(R)$:
$intend(R,know(R,bel(S,X)))$
\end{itemize}

\item Post-conditions on $S$ and $R$:
\begin{itemize}
\item $Pos(S)$: $know(S,know(R,bel(S,X)))$
 \item $Pos(R)$:
$know(R,bel(S,X))$
\end{itemize}

\item Action completion:
\begin{itemize}
\item
$know(R,bel(S,X))$
\end{itemize}
\end{itemize}
\vspace{-.75\baselineskip}}}
\caption{Semantics for \emph{tell} \protect\cite{LabrouSAKQML}.}
\label{fig:LST}
\end{figure}

\begin{figure}[!ht]
\centering \fbox{\parbox{.98\linewidth}{ \vspace{-.75\baselineskip}
\small\ttfamily
\begin{itemize}
\item
Pre-conditions on the states of $S$ and $R$:

\begin{itemize}
\item $Pre(S)$:
$want(S,know(S,Y)) \wedge know(S, intend(R, process(R,M)))$
where $Y$ is either $bel(R,X)$ or $\neg bel(R,X)$ and
$M$ is $\mbox{\textrm{\textit{ask-if}}}(S,R,X)$

\item $Pre(R)$:
$intend(R,process(R,M))$
\end{itemize}

\item
Post-conditions about $R$ and $S$:

\begin{itemize}
\item $Pos(S)$:
$intend(S,know(S,Y))$

\item $Pos(R):$
$know(R,want(S,know(S,Y)))$
\end{itemize}

\item
Action completion:

\begin{itemize}
\item
$know(S,Y)$
\end{itemize}

\end{itemize}
\vspace{-.75\baselineskip}}}
\caption{Semantics for \emph{ask-if} \protect\cite{LabrouSAKQML}.}
\label{fig:LSA}
\end{figure}

As noted above, one of the key problems with this (widely used)
approach to giving semantics to agent communication languages is that
there is no way to determine whether or not any software component
that uses such a communication language complies with the
semantics. This is because the semantics makes reference to mental
states, and we have in general no principled way to attribute such
mental states to arbitrary pieces of software.  This is true of the
semantic approaches to both KQML and FIPA, as discussed by both
Wooldridge \citeyear{Wool98} and Singh \citeyear{SinghACLRP}. As an
example, consider a legacy software component wrapped in an agent that
uses KQML or FIPA to interoperate with other agents. One cannot prove
communication properties of such system, as there is no precise
definition of when the legacy system believes that (or intends to
achieve a state of the world where) some proposition is true. Our
approach builds on the work of Bordini and Moreira
\citeyear{BordiniPBPAOPL-ATPA}, which presented a precise definition
of what it means for an \as agent to believe, desire, or intend a
certain formula; that approach is also adopted in our work on
model-checking for \as \cite{BordiniMCRA}. As a consequence, we are
able to successfully and meaningfully apply speech act-style semantics
to communication in \as. The drawback, of course, is that the approach
is, formally, limited to \as agents, even though the same ideas can
be used in work on semantics for other agent languages.

\section{Syntax and Semantics of AgentSpeak}
\label{secSSA}

The \as programming language was introduced by Rao
\citeyear{RaoASLBDIASOLCL}. It can be understood as a natural
extension of logic programming for the BDI agent architecture, and
provides an elegant abstract framework for programming BDI agents. The
BDI architecture is, in turn, perhaps one of the major approaches to
the implementation of rational practical reasoning agents
\cite{WoolRRA}.

An \as\ agent is created by the specification of a set of beliefs
forming the initial \emph{belief base} and a set of plans forming the
\emph{plan library}. An agent's belief base is a set of ground
first-order predicates, which will change over time to represent the
current state of the environment as perceived by the agent.

\as\ distinguishes two types of goals: \emph{achievement goals} and
\emph{test goals}. Achievement and test goals are predicates (as with
beliefs), prefixed with one of the operators `\texttt{!}' and
`\texttt{?}', respectively. Achievement goals state that the agent
wants to achieve a state of the world where the associated predicate
is true; in practice, as we will see, this is done by the execution of
a plan. A test goal returns a unification for the associated predicate
with one of the agent's beliefs; it fails if no such unification is
possible. A \emph{triggering event} defines which events may initiate
the execution of a plan. An \emph{event} can be internal (when a
subgoal needs to be achieved), or external (when generated from belief
updates as a result of perceiving the environment). Additionally, with
respect to the model of communication in this paper, external events
can be related to messages received from other agents. There are two
types of triggering events: those related to the \emph{addition}
(`\texttt{+}') and \emph{deletion} (`\texttt{-}') of mental attitudes
(beliefs or goals).

Plans refer to the \emph{basic actions} that an agent is able to
perform on its environment.
A plan is formed by a \emph{triggering event}, denoting the events for
which that plan is \emph{relevant}. The triggering event is followed
by a conjunction of belief literals representing a \emph{context} for
the plan. The context must be a logical consequence of the agent's
current beliefs for the plan to be \emph{applicable} --- one of the
plans that are both relevant and applicable is chosen for execution so
as to handle a particular event. The remainder of the plan is a
sequence of basic actions or \mbox{(sub-)goals} that the agent has to
achieve (or test) when the plan is executed.

\begin{figure}[t!]
\centering
\fbox{\parbox{\linewidth}{
\vspace{-.75\baselineskip}
\small\ttfamily
\begin{tabbing}
\ \\
+co\= \(\leftarrow\) \= \kill
+concert(A,V) : likes(A)\\
\> \(\leftarrow\) \> !book\_tickets(A,V).\\
\mbox{}\\
+!book\_tickets(A,V) : \(\neg\)busy(phone)\\
\> \(\leftarrow\) \> ?phone\_number(V,N); \\
\>\> call(N); \\
\>\> \(\ldots\);\\
\>\> !choose\_seats(A,V).\\
\end{tabbing}
\vspace{-.75\baselineskip}}} \caption{Examples of \as\ plans.}
\label{figEP}
\end{figure}

Figure~\ref{figEP} shows some examples of \as\ plans. The first plan
tells us that, when a concert is announced for artist \atom{A} at
venue \atom{V} (so that, from perceiving the environment, a belief
\atom{concert(A,V)} is \emph{added} to the belief base), provided that
the agent happens to like artist \atom{A}, it will have the new
achievement goal of booking tickets for that concert. The second plan
tells us that whenever this agent adopts the goal of booking tickets
for \atom{A}'s performance at \atom{V}, provided it is the case that
the telephone is not busy, it can execute a plan consisting of
retrieving from its belief base the telephone number of venue \atom{V}
(with the test goal \atom{?phone\_number(V,N)}), performing the basic
action \atom{call(N)} (assuming that making a phone call is one of the
actions that the agent is able to perform), followed by a certain
protocol for booking tickets (indicated by `$\ldots$'), which in this
case ends with the execution of a plan for choosing the seats for such
performance at that particular venue.

Next, we formally present the syntax and semantics of \as.
Note that we do not yet consider communication; we extend the
semantics to deal with communication in Section~\ref{secSCAA}.

\subsection{Abstract Syntax}

The syntax of an \as\ agent program $ag$ is defined by the grammar
below. In \as, an agent program is simply given by a set $\BELS$ of
beliefs and a set $\PLANS$ of plans. The beliefs $\BELS$ define the
initial state of the agent's belief base (i.e., the state of the
belief base when the agent starts running), and the plans $\PLANS$
form the agent's plan library.  The atomic formul\ae\ $\AT$ of the
language are predicates, where \texttt{P} is a predicate symbol and
$t_1,\ldots, t_n$ are standard terms of first order logic. A
\emph{belief} is an atomic formula $\AT$ with no variables; we use $b$
as a meta-variable for beliefs.

\[\hspace{-24pt}
\begin{array}{l@{\quad::=\quad}l@{\quad\mid\quad}l@{\quad\mid\quad}l@{\quad\mid\quad}lr}
\AGE    & \multicolumn{5}{l}{\BELS \quad \PLANS} \\
\BELS   & \multicolumn{4}{l}{b_1 \ldots b_n} & (n \geq 0) \\
\PLANS  & \multicolumn{4}{l}{p_1 \ldots p_n} & (n \geq 1) \\
p       & \multicolumn{5}{l}{\TE : \CT \SETA h} \\
\TE     & +\AT & -\AT & +g & -g \\
\CT     & \CT_1 & \multicolumn{4}{l}{\TR} \\
\CT_1   & \AT & \neg\AT & \CT_1~\wedge~\CT_1 \\
h       & h_1 \VIR \TR & \multicolumn{4}{l}{\TR} \\
h_1     & a & g & u & h_1 \VIR h_1\quad \\
\AT     & \multicolumn{4}{l}{\texttt{P}(t_1,\ldots,t_n)} & (n \geq 0) \\
\multicolumn{1}{l@{\mid~\quad}}{~} &
          \multicolumn{4}{l}{\texttt{P}(t_1, \ldots, t_n)[s_1, \ldots, s_m]}
                                                       & (n \geq 0, m > 0) \\
s       & \PERCEPT  & \SELF & \multicolumn{3}{l}{\ID} \\
a       & \multicolumn{4}{l}{\texttt{A}(t_1,\ldots,t_n)} & (n \geq 0) \\
g       & !\AT & \multicolumn{3}{l}{?\AT} \\
u       & +b & \multicolumn{3}{l}{-\AT}
\end{array}
\]

The grammar above gives an alternative definition for $\AT$, extending
the conventional syntactic form of predicates. The extension allows
``annotations'' to be associated with a predicate; this is an
extension of \as's original syntax motivated by our work on
communication, which is discussed below in Section~\ref{ssecSe}. For
the time being, suffice it to say that the idea is to annotate each
\emph{atomic formula} with its \emph{source}: either a term $\ID$
identifying which agent previously communicated that information,
$\SELF$ to denote beliefs created by the agent itself (through belief
update operations within a plan, as described below), or $\PERCEPT$ to
indicate that the belief was acquired through perception of the
environment. So, for example, if agent $i$ has a belief
$concert(a,v)[j]$ in its belief base, this would mean that agent $j$
had previously informed agent $i$ of $concert(a,v)$ --- in other
words, that $j$ wanted $i$ to believe that there will be a concert by
$a$ at $v$.  Similarly, $concert(a,v)[percept,j]$ would mean that
$a$'s concert at $v$ is believed not only because $j$ has informed
agent $i$ of this, but because $i$ itself also perceived this fact
(e.g., by seeing a poster when walking past the theatre).

A plan in \as is given by $p$ above, where $\TE$ is the
\emph{triggering event}, $\CT$ is the plan's context, and $h$ is
sequence of actions, goals, or belief updates (which should be thought
of as ``mental notes'' created by the agent itself). We refer to $\TE
: \CT$ as the \emph{head} of the plan, and $h$ is its \emph{body}. The
set of plans of an agent is given by $\PLANS$. Each plan has as part
of its head a formula $\CT$ that specifies the conditions under which
the plan can be chosen for execution.

A triggering event $\TE$ can then be the addition or the deletion of a
belief from an agent's belief base (denoted $+\AT$ and $-\AT$,
respectively), or the addition or the deletion of a goal ($+g$ and
$-g$, respectively\footnote{Triggering events of the form $-g$, in our
  approach, are used in practice for handling plan failure. Although
  we have left this construct in the grammar, we have omitted the
  discussion and formalisation of plan failure for clarity, as the
  focus in this paper is on the semantics of communication.}). For
plan bodies, we assume the agent has at its disposal a set of
\emph{actions} and we use $a$ as a meta-variable ranging over them. We
are largely unconcerned here with respect to exactly what such actions
are. Actions are written using the same notation as predicates, except
that an action symbol \texttt{A} is used instead of a predicate
symbol. Goals $g$ can be either \emph{achievement goals} ($!\AT$) or
\emph{test goals} ($?\AT$). Finally, $+b$ and $-\AT$ (in the body of a
plan) represent operations for updating ($u$) the belief base by,
respectively, adding or removing beliefs; recall that an atomic
formula must be ground if it is to be added to the belief base.

\subsection{Semantics}
\label{ssecSe}

We define the semantics of \as using operational semantics, a widely
used method for giving semantics to programming
languages \cite{Plotkin:SOS}. The operational semantics is given by a
set of rules that define a transition relation between configurations
\CFG{s} where:

\begin{itemize}

\item An agent program $\AGE$ is, as defined above, a set of beliefs
  $\BELS$ and a set of plans $\PLANS$.

\item An agent's circumstance $C$ is a tuple $\langle I, E, A \rangle$ where:

\begin{itemize}
\item $I$ is a set of \emph{intentions} $\{i, i', \ldots\}$. Each intention
  $i$ is a stack of partially instantiated plans.

\item $E$ is a set of \emph{events} $\{ (\TE,i), (\TE', i'), \ldots \}$. Each
  event is a pair $(\TE,i)$, where $\TE$ is a triggering event and $i$ is an
  intention --- a stack of plans in case of an internal event, or the empty
  intention $\TR$ in case of an external event.
  When the belief revision function (which is not part of the \as
  interpreter but rather of the agent's overall architecture), updates
  the belief base, the associated events --- i.e., additions and
  deletions of beliefs --- are included in this set. These are called
  \emph{external} events; internal events are generated by additions
  or deletions of goals from plans currently executing.

\item $A$ is a set of \emph{actions} to be performed in the
  environment.
\end{itemize}

\item $M$ is a tuple $\langle In, Out, SI \rangle$ whose components
  characterise the following aspects of communicating agents (note
  that communication is asynchronous):

\begin{itemize}
  
\item $In$ is the mail inbox: the system includes all messages
  addressed to this agent in this set. Elements of this set have the
  form $\msg{\MID,\ID,\ILF,\CNT}$, where $\MID$ is a message
  identifier, $\ID$ identifies the sender of the message, $\ILF$ is
  the illocutionary force of the message, and $\CNT$ its content: a
  (possibly singleton) set of AgentSpeak predicates or plans,
  depending on the illocutionary force of the message.

\item $Out$ is where the agent posts messages it wishes to send; it is
  assumed that some underlying communication infrastructure handles
  the delivery of such messages. (We are not concerned with this
  infrastructure here.) Messages in this set have exactly the same
  format as above, except that here $\ID$ refers to the agent to which
  the message is to be sent.

\item $SI$ is used to keep track of intentions that were suspended due
  to the processing of communication messages; this is explained in
  more detail in the next section, but the intuition is as follows:
  intentions associated with illocutionary forces that require a reply
  from the interlocutor are suspended, and they are only resumed when
  such reply has been received.

\end{itemize}

\item It is useful to have a structure which keeps track of temporary
  information that may be subsequently required within a reasoning
  cycle. $T$ is a tuple $\langle R, \mathit{Ap}, \iota, \varepsilon,
  \rho \rangle$ with such temporary information; these components are
  as follows:

\begin{itemize}
\item
  $R$ is the set of \emph{relevant plans} (for the event being handled).

\item
  $\mathit{Ap}$ is the set of \emph{applicable plans} (the relevant plans
  whose contexts are true).

\item $\iota$, $\varepsilon$, and $\rho$ record a particular intention, event,
  and applicable plan (respectively) being considered along the execution of
  one reasoning cycle.
\end{itemize}

\item The current step within an agent's reasoning cycle is
  symbolically annotated by $s \in \{ \ProcMsg, \SelEv, \RelPl,
  \ApplPl, \SelAppl, \AddIM, \SelInt, \ExecInt, \ClrInt \}$. These
  labels stand for, respectively: processing a message from the
  agent's mail inbox, selecting an event from the set of events,
  retrieving all relevant plans, checking which of those are
  applicable, selecting one particular applicable plan (the intended
  means), adding the new intended means to the set of intentions,
  selecting an intention, executing the selected intention, and
  clearing an intention or intended means that may have finished in
  the previous step.
\end{itemize}

In the interests of readability, we adopt the following notational
conventions in our semantic rules:

\begin{itemize}

\item If $C$ is an AgentSpeak agent circumstance, we write $C_E$ to
  make reference to the $E$ component of $C$, and similarly for other
  components of a configuration.

\item We write $T_{\iota} = \_$ (the underscore symbol) to indicate
  that there is no intention presently being considered in that
  reasoning cycle.  Similarly for $T_\rho$ and $T_\varepsilon$.

\item We write $i[p]$ to denote the intention that has plan $p$ on top
  of intention $i$.

\end{itemize}

The \as\ interpreter makes use of three \emph{selection functions}
that are defined by the agent programmer. The selection function $\SE$
selects an event from the set of events $\CE$; the selection function
$\SAP$ selects one applicable plan given a set of applicable plans;
and $\SI$ selects an intention from the set of intentions $\CI$ (the
chosen intention is then executed). Formally, all the selection
functions an agent uses are also part of its configuration (as is the
social acceptance function that we mention later when we formalise
agent communication). However, as they are defined by the agent
programmer at design time and do not (in principle) change at run
time, we avoid including them in the configuration for the sake of
readability.

We define some functions which help simplify the semantics. If $p$ is
a plan of the form $\TE : \CT \SETA h$, we define $\TREV(p) = \TE$ and
$\CTXT(p) = \CT$. That is, these projection functions return the
triggering event and the context of the plan, respectively. The
$\TREV$ function can also be applied to the head of a plan rather than
the whole plan, but works similarly in that case.

Next, we need to define the specific (limited) notion of logical
consequence used here.  We assume a procedure that computes the most
general unifier of two literals (as usual in logic programming), and
with this, define the logical consequence relation $\models$ that is
used in the definitions of the functions for checking for relevant and
applicable plans, as well as executing test goals. Given that we have
extended the syntax of atomic formul\ae\ so as to include annotations
of the sources for the information symbolically represented by it, we
also need to define $\models$ in our particular context, as follows.

\begin{defn}
\label{def:lc} We say that an atomic formula $\AT_1$ with
annotations ${s_1}_1,\ldots,{s_1}_n$ is a logical consequence of a
set of ground atomic formul\ae\ $\BELS$, written $\BELS \models
\AT_1[{s_1}_1,\ldots,{s_1}_n]$ if, and only if, there exists
$\AT_2[{s_2}_1,\ldots,{s_2}_m] \in \BELS$ such that (i) $\AT_1\theta
= \AT_2$, for some most general unifier $\theta$, and (ii)
$\{{s_1}_1,\ldots,{s_1}_n\} \subseteq \{{s_2}_1,\ldots,{s_2}_m\}$.
\end{defn}

The intuition is that, not only should predicate $\AT$ unify with some
belief in $\BELS$ (i), but also that all specified sources of
information for $\AT$ should be corroborated in $\BELS$ (ii). Thus,
for example, $\atom{p(X)[ag$_1$]}$ follows from
$\{\atom{p(t)[ag$_1$,ag$_2$]}\}$, but $\atom{p(X)[ag$_1$,ag$_2$]}$
does \emph{not} follow from $\{\atom{p(t)[ag$_1$]}\}$.  More
concretely, if, in order to be applicable, a plan requires that a
drowning person was explicitly perceived rather than communicated by
another agent (which can be represented by
$\atom{drowning(Person)[percept]}$), this follows from a belief
$\atom{drowning(man)[percept,passerby]}$ (i.e., that this was both
perceived and communicated by a passerby). On the other hand, if the
required context was that two independent sources provided the
information, say $\atom{cheating(Person)[witness1,witness2]}$, this
cannot be inferred from a belief $\atom{cheating(husband)[witness1]}$.

In order to make some semantic rules more readable, we use two
operations on a belief base (i.e., a set of annotated ground atomic
formul\ae). We use $\BELS'=\BELS+b$ to say that $\BELS'$ is as $\BELS$
except that $\BELS' \models b$. Similarly $\BELS'=\BELS-b$ means that
$\BELS'$ is as $\BELS$ except that $\BELS' \not\models b$.

A plan is considered \emph{relevant} in relation to a triggering event
if it has been written to deal with that event. In practice, this is
checked by trying to unify the triggering event part of the plan with
the triggering event within the event that has been selected for
treatment in that reasoning cycle. In the definition below, we use the
logical consequence relation defined above to check if a plan's
triggering event unifies with the event that has occurred.  To do
this, we need to extend the $\models$ relation so that it also applies
to triggering events instead of predicates. In fact, for the purposes
here, we can consider that any operators in a triggering event (such
as `$+$' or `$!$') are part of the predicate symbol or, more
precisely, let $\AT_1$ be the predicate (with annotation) within
triggering event $\TE_1$ and $\AT_2$ the one within $\TE_2$, then
$\{\TE_2\} \models \TE_1$ if, and only if, $\{\AT_2\} \models \AT_1$
and, of course, the operators prefixing $\TE_1$ and $\TE_2$ are
exactly the same. Because of the requirement of inclusion of
annotations, the converse might not be true.

\begin{defn}
  \label{def:rel} Given plans $\PLANS$ of an agent and a triggering
  event $\TE$, the set $\RELPLANS(\PLANS, \TE)$ of relevant plans for
  $\TE$ is defined as follows:
\[
\RELPLANS(\PLANS, \TE) = \{(p, \theta) ~|~ p\in\PLANS \mbox{\textit{
    and }} \theta \mbox{\textit{ is s.t.\ }} \{\TE\} \models
\TREV(p)\theta \}.
\]
\end{defn}

The intuition regarding annotations is as follows. The programmer
should include in the annotations of a plan's triggering event all the
sources that must have generated the event for that plan to be
relevant (or include no annotation if the source of information is not
important for the plan to be considered relevant). For the plan to be
relevant, it therefore suffices for the annotations in the plan's
triggering event to be a subset of those in the event that occurred. A
plan with triggering event $\mathtt{+!p(X)[s]}$ is relevant for an
event $\langle\mathtt{+!p(t)[s,t]},\TR\rangle$ since $\RELPLANS$
requires that $\{\mathtt{p(t)[s,t]}\} \models \mathtt{p(X)[s]}\theta$
(for some most general unifier $\theta$), which in turn requires that
$\{\mathtt{s}\} \subseteq \{\mathtt{s},\mathtt{t}\}$. As a
consequence, for a plan with a triggering event that has no
annotations (e.g., $\mathtt{+!p(X)}$) to be relevant for a particular
event (say, $\langle\mathtt{+!p(t)[ag_1]},i\rangle$) it only requires
that the predicates unify in the usual sense since $\{\} \subseteq S$,
for any set $S$.

A plan is \emph{applicable} if it is relevant and its context is a
logical consequence of the agent's beliefs. Again we need to extend
slightly the definition of $\models$ given above. A plan's context is
a conjunction of literals ($\LIT$ is either $\AT$ or $\neg\AT$). We
can say that $\BELS \models \LIT_1 \wedge \ldots \wedge \LIT_n$ if,
and only if, $\BELS \models \LIT_i$ if $\LIT_i$ is of the form $\AT$,
and $\BELS \not\models \LIT_i$ of $\LIT_i$ is of the form $\neg\AT$,
for $1 \leq i \leq n$. The function for determining the applicable
plans in a set of relevant plans is formalised as follows.

\begin{defn}
\label{def:applic} Given a set of relevant plans $R$ and the
beliefs $\BELS$ of an agent, the set of applicable plans
$\APPLPLANS(\BELS, R)$ is defined as follows:
\[
\APPLPLANS(\BELS, R) = \{ (p,\theta' \circ \theta) ~|~ (p,\theta)\in R
\mbox{\textit{ and }} \theta'
\mbox{\textit{ is s.t.\ }} \BELS \models \CTXT(p)\theta\theta' \}.
\]
\end{defn}

We need another function to be used in the semantic rule for when the
agent is to execute a test goal. The evaluation of a test goal $?\AT$
consists in testing if the formula $\AT$ is a logical consequence of
the agent's beliefs. The function returns a set of most general
unifiers all of which make the formula $\AT$ a logical consequence of
a set of formul\ae\ $\BELS$, as follows.

\begin{defn}
\label{def:test} Given a set of formul\ae\ $\BELS$ and a formula
$\AT$, the set of substitutions $\TEST(\BELS, \AT)$ produced
by testing $\AT$ against $\BELS$ is defined as follows:

\[
\TEST(\BELS,\AT) = \{ \theta ~|~ \BELS \models \AT\theta \}.
\]
\end{defn}

Next, we present the reasoning cycle of \as agents and the rules which
define the operational semantics.

\subsection{Reasoning Cycle}
\label{sec:reasoningcycle}

\begin{figure}[thb]
\begin{center}
\fbox{\includegraphics[width=.98\linewidth]{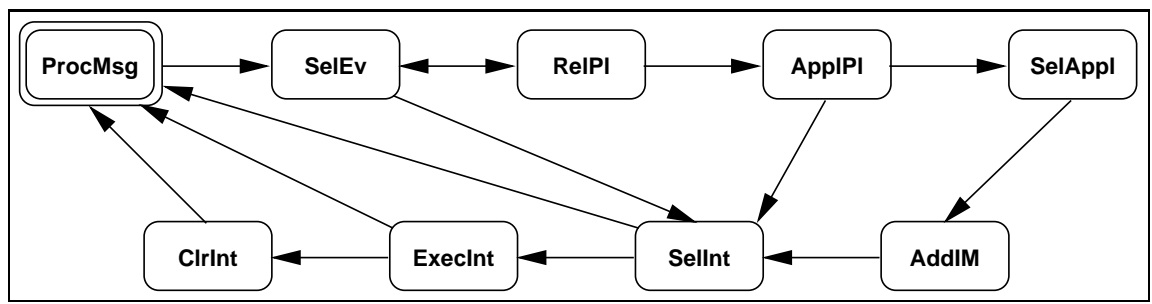}}
\caption{The \as agent reasoning cycle.}
\label{fig:cycle}
\end{center}
\end{figure}

Figure~\ref{fig:cycle} shows the possible transitions between the
various steps in an agent's reasoning cycle as determined by an \as
interpreter. The labels in the nodes identify each step of the cycle,
which are: processing received messages ($\ProcMsg$); selecting an
event from the set of events ($\SelEv$); retrieving all relevant plans
($\RelPl$); checking which of those are applicable ($\ApplPl$);
selecting one particular applicable plan (the intended means)
($\SelAppl$); adding the new intended means to the set of intentions
($\AddIM$); selecting an intention ($\SelInt$); executing the selected
intention ($\ExecInt$), and clearing an intention or intended means
that may have finished in the previous step ($\ClrInt$).

In the general case, an agent's initial configuration is
\CFG{\ProcMsg}, where $\AGE$ is as given by the agent program, and all
components of $C$, $M$, and $T$ are empty. Note that a reasoning cycle
starts with processing received messages ($\ProcMsg$) --- the
semantics for this part of the reasoning cycle are given in the main
section of this paper. After that, the original \as reasoning cycle
takes place. An event selection ($\SelEv$) is made, which is followed
by determining relevant and applicable plans ($\RelPl$ and $\ApplPl$,
respectively). One of the relevant plans is then selected
($\SelAppl$); note that when there are no events to be treated or when
there are no applicable plans to deal with an event the agent turns
its attention to the selection of an intended means ($\SelInt$) to be
executed next. After one of the relevant plans is selected
($\SelAppl$) and an instance of that plan becomes an ``intended
means'' and is therefore included in the set of intentions ($\AddIM$).
When there are more than one intention (which is normally the case
except for extremely simple agents), one of those intentions is
selected ($\SelInt$) and executed ($\ExecInt$).

These are the most important transitions; the others will be made
clearer when the semantics is presented. The rules which define the
transition systems giving operational semantics to \as (without
communication) are presented next.

\subsection{Semantic Rules}
\label{appSR}

In this section, we present an operational semantics for \as that
formalises the transitions between possible steps of the
interpretation of \as agents as shown in Figure~\ref{fig:cycle}. In
the general case, an agent's initial configuration is \CFG{\ProcMsg},
where $\AGE$ is as given by the agent program, and all components of
$C$, $M$, and $T$ are empty. Note that a reasoning cycle starts with
processing received messages ($\ProcMsg$), according to the most
recent extension of the semantics to be presented in
Section~\ref{secSCAA}. An event selection ($\SelEv$) is then made,
starting the reasoning cycle as originally defined for the language,
which is the part of the semantics presented below.

\begin{paragraph*}{Event Selection:}
  The rule below assumes the existence of a selection function $\SE$
  that selects events from a set of events $E$. The selected event is
  removed from $E$ and it is assigned to the $\varepsilon$ component
  of the temporary information. Rule \rname{SelEv$_2$} skips to the
  intention execution part of the cycle, in case there are no events
  to handle.

\infrule[SelEv$_1$]
{\label{rule:SelEv1}
     \SE(\CE) = \langle\TE, i\rangle }
{
     \CFG{\SelEv} \trans \CFGctp{\RelPl} \\[\rwsep]
\begin{array}{llcl}
\mbox{\emph{where:} } & \CEli   & = & \CE \setminus \{\langle\TE, i\rangle\} \\
                      & \TEPSli & = & \langle\TE, i\rangle
\end{array}
}

\infrule[SelEv$_2$]
{\label{rule:SelEv2}
     \CE = \VAZ
}
{
     \CFG{\SelEv} \trans \CFG{\SelInt}
}
\end{paragraph*}

\begin{paragraph*}{Relevant Plans:}
  Rule \rname{Rel$_1$} assigns the set of relevant plans to component
  $\TRE$.  Rule \rname{Rel$_2$} deals with the possibility that there
  are no relevant plans for an event, in which case the event is
  simply discarded. In fact, an intention associated with the event
  might also be discarded: if there are no relevant plans to handle an
  event generated by that intention, it cannot be further executed. In
  practice, instead of simply discarding the event (and possibly an
  intention with it), this leads to the activation of the plan failure
  mechanism, which we do not discuss here for clarity of presentation,
  as discussed earlier.

\infrule[Rel$_1$] { \TEPS=\langle\TE,i\rangle \andalso \RELPLANS(\AGPLANS,\TE)
  \neq \VAZ }
{     \CFG{\RelPl} \trans \CFGtp{\ApplPl} \\[\rwsep]
\begin{array}{llcl}
 \mbox{\emph{where:} } & \TREli & = & \RELPLANS(\AGPLANS, \TE)
\end{array}
}

\infrule[Rel$_2$]
{
   \RELPLANS(\AGPLANS, \TE) = \VAZ
}
{
    \CFG{\RelPl} \trans \CFG{\SelEv} \\[\rwsep]
}

An alternative approach for situations where there are no relevant
plans for an event was introduced by \citeauthor{AnconaCooAScape}
\citeyear{AnconaCooAScape}. It assumes that in some cases, explicitly
specified by the programmer, the agent will want to ask other agents
what are the recipes they use for handling such events. The mechanism
for plan exchange between AgentSpeak agents they proposed allows the
programmer to specify which triggering events should generate attempts
to retrieve external plans, which plans an agent agrees to share with
others, what to do once the plan has been used for handling that
particular event instance, and so forth.
\end{paragraph*}

\begin{paragraph*}{Applicable Plans:}
  The rule \rname{Appl$_1$} assigns the set of applicable plans to the
  $\TAP$ component; rule \rname{Appl$_2$} applies when there are no
  applicable plans for an event, in which case the event is simply
  discarded. Again, in practice, this normally leads to the plan
  failure mechanism being activated, rather than simply discarding the
  event (and the whole intention with it).

\infrule[Appl$_1$]
{
    \APPLPLANS(\AGBELS, \TRE) \neq \VAZ
}
{
    \CFG{\ApplPl} \trans \CFGtp{\SelAppl} \\[\rwsep]
\begin{array}{llcl}
\mbox{\emph{where:} }
                     & \TAPli & = & \APPLPLANS(\AGBELS, \TRE)
\end{array}
}

\infrule[Appl$_2$] {
   \APPLPLANS(\AGBELS, \TRE) = \VAZ
}
{
    \CFG{\ApplPl} \trans \CFG{\SelInt} \\[\rwsep]
}

\end{paragraph*}

\begin{paragraph*}{Selection of an Applicable Plan:}
  This rule assumes the existence of a selection function $\SAP$ that
  selects one plan from a set of applicable plans $\TAP$. The selected
  plan is then assigned to the $\TRHO$ component of the configuration.

\infrule[SelAppl] {
    \SAP(\TAP) = (p, \theta)
}
{
    \CFG{\SelAppl} \trans \CFGtp{\AddIM} \\[\rwsep]
\begin{array}{llcl}
\mbox{\emph{where:} } & \TRHOli & = & (p,\theta)
\end{array}
}
\end{paragraph*}

\begin{paragraph*}{Adding an Intended Means to the Set of Intentions:}
  Events can be classified as external or internal (depending on
  whether they were generated from the agent's perception, or whether
  they were generated by the previous execution of other plans,
  respectively). Rule \rname{ExtEv} determines that, if the event
  $\varepsilon$ is external (which is indicated by $\TR$ in the
  intention associated to $\varepsilon$), a new intention is created
  and the only intended means in that new intention is the plan $p$
  assigned to the $\rho$ component. If the event is internal, rule
  \rname{IntEv} determines that the plan in $\rho$ should be put on
  top of the intention associated with the event.

\infrule[ExtEv] {
    \TEPS=\langle\TE,\TR\rangle \andalso \TRHO = (p,\theta)
}
{
    \CFG{\AddIM} \trans \CFGcp{\SelInt} \\[\rwsep]
\begin{array}{llcl}
\mbox{\emph{where:} } & \CIli & = & \CI \cup \{~[p\theta]~\}
\end{array}
}

\infrule[IntEv] {
    \TEPS=\langle\TE,i\rangle \quad \TRHO = (p,\theta)
}
{
    \CFG{\AddIM} \trans \CFGcp{\SelInt} \\[10pt]
\begin{array}{llcl}
\mbox{\emph{where:} } & \CIli    & = & \CI \cup \{~i[(p\theta)]~\} \\
\end{array}
}

Note that, in rule \rname{IntEv}, the whole intention $i$ that
generated the internal event needs to be inserted back in $\CI$, with
$p$ pushed onto the top of that intention. This is related to resuming
\emph{suspended intentions}; the suspending of intentions appears in
rule \rname{AchvGl} below.
\end{paragraph*}

\begin{paragraph*}{Intention Selection:}
  Rule \rname{SelInt$_1$} assumes the existence of a function $\SI$
  that selects an intention for processing next, while rule
  \rname{SelInt$_2$} takes care of the situation where the set of
  intentions is empty (in which case the reasoning cycle simply starts
  again).

\infrule[SelInt$_1$]
{
    \CI \neq \{  \} \qquad \SI(\CI) = i
}
{
    \CFG{\SelInt} \trans \CFGtp{\ExecInt} \\[\rwsep]
\begin{array}{llcl}
 \mbox{\emph{where:} } & \TIOTAli & = & i
\end{array}
}
\infrule[SelInt$_2$]
{
    \CI = \{  \}
}
{
    \CFG{\SelInt} \trans \CFG{\ProcMsg}
}

\end{paragraph*}

\begin{paragraph*}{Executing an Intention:}
  The group of rules below express the effects of executing a formula
  in the body of the plan. Each rule deals with one type of formula
  that can appear in a plan body. Recall from Section~\ref{ssecSe}
  that an intention is a stack of (partially instantiated) plan
  instances; a plan instance is a copy of a plan from the agent's plan
  library). The plan instance to be executed is always the one at the
  top of the intention that was selected in the previous step (rule
  \rname{SelInt$_1$}); the specific formula to be executed is the one
  at the beginning of the body of that plan.

  \noindent\emph{Actions:} When the formula to be executed is an
  action, the action $a$ in the body of the plan is added to the set
  of actions $A$ (which, recall, denotes that the action is to be
  executed using the agent's effectors). The action is removed from
  the body of the plan and the intention is updated to reflect this
  removal.

\infrule[Action] {
    \TIOTA = i[\HEAD \SETA a \VIR h]
}
{
    \CFG{\ExecInt} \trans \CFGcp{\ClrInt} \\[\rwsep]
\begin{array}{llcl} \mbox{\emph{where:} }
                  & \CAli    & = & \CA \cup \{a\}\\
                  & \CIli    & = & (\CI \setminus \{ \TIOTA \}) \cup
                                   \{ i[\HEAD \SETA h] \}
\end{array}
}

\noindent\emph{Achievement Goals:} 
This rule registers a new internal event in the set of events $E$.
This event can then be selected for handling in a future reasoning
cycle (see rule \rname{SelEv$_1$}). When the formula being executed is
a goal, the formula is not removed from the body of the plan, as in
the other cases. This only happens when the plan used for achieving
that goal finishes successfully; see rule~\rname{ClrInt$_2$}. The
reasons for this are related to further instantiation of the plan
variables as well as handling plan failure.

\infrule[AchvGl]
{
    \TIOTA = i[\HEAD \SETA\ !\AT \VIR h]
}
{
    \CFG{\ExecInt} \trans \CFGcp{\ProcMsg} \\[\rwsep]
\begin{array}{llcl}
\mbox{\emph{where:} }
                & \CEli    & = & \CE \cup \{ \langle +!\AT, \TIOTA \rangle \}\\
                & \CIli    & = & \CI \setminus \{ \TIOTA \}
\end{array}
}

Note how the intention that generated the internal event is removed
from the set of intentions $\CI$, capturing the idea of
\emph{suspended intentions}.  In a plan body, if we have `$!g ; f$'
(where $f$ is any formula that can appear in plan bodies), this means
that, before $f$ can be executed, the state of affairs represented by
goal $g$ needs to be achieved (through the execution of some relevant,
applicable plan). The goal included in the new event created by rule
\rname{AchvGl} above is treated as any other event, which means it
will go the set of events until it is eventually selected in a later
reasoning cycle, according to the agent's specific priorities for
selecting events (rule \rname{SelEv$_1$}). Meanwhile, that plan (with
formula $f$ to be executed next) can no longer be executed, hence the
whole intention is suspended by being placed, within the newly created
event, in the set of events and removed from the set of intentions.
When the event created by the rule above is selected and an applicable
plan for achieving $g$ has been chosen, that intended means is pushed
on top of the suspended intention, which can then be \emph{resumed}
(i.e., moved back to the set of intentions), according to rule
\rname{IntEv}. The next time that intention is selected, its execution
will then proceed with a plan for achieving $g$ at the top, and only
when that plan is finished will $f$ be executed (as that plan, now
without the achieved goal, will be at the top of the intention again);
further details on suspended intentions can be found in the \as
literature \cite<e.g., see>{BordiniPBPAOPL-ATPA}.

\noindent\emph{Test Goals:} These rules are used when a test goal
formula $?\AT$ is to be executed. Rule \rname{TestGl$_1$} is used when
there is a set of substitutions that can make $\AT$ a logical
consequence of the agent's beliefs, which means that the test goal
succeeded. If the test goal succeeds, the substitution is applied to
the whole intended means, and the reasoning cycle can be continued. If
that is not the case, it might turn out that the test goal is used as
a triggering event of a plan, which is used by programmers to
formulate more sophisticated queries.  Rule \rname{TestGl$_2$} is used
in such case: it generates an internal event, which may trigger the
execution of a plan, as for achievement goals. If to carry out a plan
an agent is required to obtain information (at the time of actual
execution of the plan) which is not directly available in its belief
base, a plan for a test goal can be written which, for example, sends
messages to other agents, or processes available data, so that the
particular test goal can be concluded (producing an appropriate
instantiation of logical variables). If an internal event is generated
for the test goal being executed, the process is very similar to
achievement goals, where the intention is suspended until a plan is
selected to achieve the goal, as explained above.

\infrule[TestGl$_1$]
{
    \TIOTA = i[\HEAD \SETA\ ?\AT \VIR h] \andalso
    \TEST(\AGBELS, \AT) \neq \VAZ
}
{
    \CFG{\ExecInt} \trans \CFGcp{\ClrInt} \\[\rwsep]
\begin{array}{llcl}
\mbox{\emph{where:} }
                     & \CIli    & = & (\CI \setminus \{\TIOTA\}) \cup
                                      \{i[(\HEAD \SETA h)\theta]\} \\
                     &          &   &   \quad  \theta\in\TEST(\AGBELS, \AT)
\end{array}
}

\infrule[TestGl$_2$]
{
    \TIOTA = i[\HEAD \SETA\ ?\AT \VIR h] \andalso
    \TEST(\AGBELS, \AT) = \VAZ
}
{
    \CFG{\ExecInt} \trans \CFGcp{\ClrInt} \\[\rwsep]
\begin{array}{llcl}
\mbox{\emph{where:} }
                & \CEli    & = & \CE \cup
                           \{ \langle +?\AT,\TIOTA \rangle \}\\
                & \CIli    & = & \CI \setminus \{ \TIOTA \} \\
\end{array}
}

\noindent\emph{Updating Beliefs:} In the rules below, the set of
beliefs of the agent is modified in a way that either an atomic
formula (with annotation $\SELF$) is included in the new set of
beliefs (rule \rname{AddBel}) or it is removed from there (rule
\rname{DelBel}). Both rules add a new event to the set of events $E$,
and update the intention by removing from it the $+b$ or $-\AT$
formula just executed. Note that belief deletions can have variables
($\AT$), whilst only ground atoms ($b$) can be added to the belief
base.

\infrule[AddBel] {
    \TIOTA = i[\HEAD \SETA +b \VIR h]
}
{
    \CFG{\ExecInt} \trans \CFGacp{\ClrInt} \\[\rwsep]
\begin{array}{llcl}
\mbox{\emph{where:} }
  & \AGE'_{\BELS}  & = & \AGE_{\BELS} + b[\SELF]\\
          & \CEli  & = & \CE \cup \{ \langle +b[\SELF], \TR \rangle \} \\
          & \CIli  & = & (\CI \setminus \{\TIOTA\}) \cup \{i[\HEAD \SETA h]\}
\end{array}
}

\infrule[DelBel]
{
    \TIOTA = i[\HEAD \SETA -\AT \VIR h]
}
{
    \CFG{\ExecInt} \trans \CFGacp{\ClrInt} \\[\rwsep]
\begin{array}{llcl}
\mbox{\emph{where:} }
  & \AGE'_{\BELS}  & = & \AGE_{\BELS} - \AT[\SELF] \\
          & \CEli  & = & \CE \cup \{ \langle -\AT[\SELF], \TR \rangle \} \\
          & \CIli  & = & (\CI \setminus \{\TIOTA\}) \cup \{i[\HEAD \SETA h]\}
\end{array}
}
\end{paragraph*}

\begin{paragraph*}{Clearing Intentions:}
  Finally, the following rules remove empty intended means or
  intentions from the set of intentions. Rule \rname{ClrInt$_1$}
  simply removes a whole intention when there is nothing else to be
  executed in that intention. Rule \rname{ClrInt$_2$} clears the
  remainder of the plan with an empty body currently at the top of a
  (non empty) intention. In this case, it is necessary to further
  instantiate the plan below the finished plan (currently at the top
  of that intention), and remove the goal that was left at the
  beginning of the body of the plan below (see rules \rname{AchvGl}
  and \rname{TestGl}). Note that, in this case, further ``clearing''
  might be necessary, hence the next step is still \ClrInt. Rule
  \rname{ClrInt$_3$} takes care of the situation where no (further)
  clearing is required, so a new reasoning cycle can start (at step
  \ProcMsg).

\infrule[ClrInt$_1$]
{
    j = [\HEAD \SETA \TR] \mbox{, for some } j \in \CI
}
{
    \CFG{\ClrInt} \trans \CFGcp{\ProcMsg} \\[\rwsep]
\begin{array}{llcl}
\mbox{\emph{where:} } & \CIli    & = & \CI \setminus \{j\}
\end{array}
}

\infrule[ClrInt$_2$]
{
    j = i[\HEAD \SETA \TR] \mbox{, for some } j \in \CI
}
{
    \CFG{\ClrInt} \trans \CFGcp{\ClrInt} \\[\rwsep]
\begin{array}{llcl}
\mbox{\emph{where:} } & \CIli    & = & (\CI \setminus \{ j \}) \cup
                                       \{k[(\HEAD' \SETA h)\theta]\} \\
\multicolumn{4}{l}{\mbox{\emph{if} } i = k[\HEAD' \SETA g \VIR h ]
\mbox{\emph{ and } } \theta \mbox{\emph{ is s.t. }} g\theta = \TREV(\HEAD)}
\end{array}
}

\infrule[ClrInt$_3$]
{
    j \neq [\HEAD \SETA \TR] \wedge j \neq i[\HEAD \SETA \TR]
    \mbox{, for \emph{any} } j \in \CI
}
{
    \CFG{\ClrInt} \trans \CFG{\ProcMsg}
}
\end{paragraph*}

\section{Semantics of Communicating AgentSpeak Agents}
\label{secSCAA}

The rules in the previous section give semantics to the key internal
decision making and control aspects of AgentSpeak. Furthermore, the
overall agent architecture will have sensors (with an associated
belief revision function) and effectors, in addition to an AgentSpeak
interpreter. The relation of these components to the AgentSpeak
interpreter is not essential for giving a semantics to the language
itself. It suffices to note that belief revision from perception of
the environment adds (external) events to the set $\CE$ (which is then
used in the AgentSpeak interpretation cycle), while the effectors
simply execute every action that is included by the reasoner in the
set $\CA$.

Similarly, the mechanism that allows messages to be exchanged is part
of the overall agent architecture --- it is not part of its practical
reasoning component, which is specifically what we program with \as.
The notion of internal actions in \as \cite{BordiniASXLeisbadtts} is
appropriate here: sending a message corresponds to executing the
(predefined) internal action $\SEND$ that appears in a plan body. The
underlying agent architecture ensures that the necessary technical
means is used for the message to reach the agent to which the message
is addressed. However, as we will be referring to a special type of
communication action that involves suspending intentions, we now need
to include such details in the semantics.\footnote{Some aspects of the
  whole framework are still not included in the formalisation given in
  this paper.  We extend the semantics only to the point required for
  accounting for the semantics of speech-act based messages received
  by an agent.}

The format of messages is $\langle\MID,\ID,\ILF,\CNT\rangle$, where
$\MID$ uniquely identifies the message, $\ID$ identifies the agent to
which the message is addressed (when the message is being sent) or the
agent that has sent the message (when the message is being received),
$\ILF$ is the illocutionary force (i.e., the performative) associated
with the message, and $\CNT$ is the message content. Depending on the
illocutionary force of the message, its content can be: an atomic
formula ($\AT$); a set of formul\ae\ ($\SAT$); a ground atomic formula
($b$); a set of ground atomic formul\ae\ ($\SB$); or a set of plans
($\SPL$).

A mechanism for receiving and sending messages asynchronously is then
defined. Messages are stored in a mail box and one of them is
processed by the agent at the beginning of a reasoning cycle. Recall
that, in a configuration of the transition system, $\MI$ is the set of
messages that the agent has received but has not processed yet, $\MO$
is the set of messages to be sent to other agents, and $\MSI$ is a set
of suspended intentions awaiting replies for (information request)
messages previously sent. More specifically, $\MSI$ is a set of pairs
of the form $(\MID,i)$, where $\MID$ is a message identifier that
uniquely identifies the previously sent message that caused intention
$i$ to be suspended.

When sending messages with illocutionary forces related to information
requests, we have chosen a semantics in which the intention is
suspended until a reply is received from the interlocutor, very much
in the way that intentions get suspended when they are waiting for an
internal event to be handled. With this particular semantics for
``ask'' messages, the programmer knows with certainty that any
subsequent action in the body of a plan is only executed after the
requested information has already been received. However, note that
the information received as a reply is stored directly in the agent's
belief base, so a test goal is required if the information is to be
used in the remainder of the plan.

We now give two rules for executing the (internal) action of sending a
message to another agent: the first is for the ``ask'' messages which
require suspending intentions and the second is for other types of
messages. These rules have priority over \rname{Action}; although
\rname{Action} could also be applied on the same configurations, we
assume the rules below are used if the formula to be executed is
specifically a $\SEND$ action. (We did not include this as a proviso
in rule \rname{Action} to improve readability.)

\infrule[ExecActSndAsk] {
    \TIOTA = i[\HEAD \SETA \SEND(\ID,\ILF,\CNT) \VIR h] \\
    \ILF \in \{\ASKIF,\ASKALL,\ASKHOW\}
}
{
    \CFG{\ExecInt} \trans \CFGcmp{\ProcMsg} \\[\rwsep]
\begin{array}{llcl}
\mbox{\emph{where:} } & \MOli & = & \MO \cup \{ \msg{\MID,\ID,\ILF,\CNT} \} \\
                      & \MSIli & = & \MSI \cup \{ (\MID,i[\HEAD \SETA h]) \}, \\
                      & & & \mbox{with $\MID$ a new  message identifier;}\\
                      & \CIli & = & (\CI \setminus \{ \TIOTA \})
\end{array}
}

The semantics of sending other types of illocutionary forces is then
simply to add a well-formed message to the agent's mail outbox (rule
\rname{ExecActSnd}). Note that in the rule above, as the intention is
suspended, the next step in the reasoning cycle is \ProcMsg (i.e., a
new cycle is started), whereas in the rule below it is \ClrInt, as the
updated intention --- with the sending action removed from the plan
body --- might require ``clearing'', as with any of the intention
execution rules seen in the previous section.

\infrule[ExecActSnd] {
    \TIOTA = i[\HEAD \SETA \SEND(\ID,\ILF,\CNT) \VIR h] \\
    \ILF \not\in \{\ASKIF,\ASKALL,\ASKHOW\}
}
{
    \CFG{\ExecInt} \trans \CFGcmp{\ClrInt} \\[\rwsep]
\begin{array}{llcl}
\mbox{\emph{where:} } & \MOli & = & \MO \cup \{ \msg{\MID,\ID,\ILF,\CNT} \}, \\
                      &       &   & \mbox{with $\MID$ a new message
                      identifier;} \\
                      & \CIli & = & (\CI \setminus \{ \TIOTA \})\cup \{ i[\HEAD \SETA h] \}
\end{array}
}

Whenever new messages are sent, we assume the system creates unique
\emph{message identifiers} (\MID). Later, we shall see that, when
replying to a message, the same message identifier is kept in the
message, similar to the way that \texttt{reply-with} is used in KQML.
Thus, the receiving agent is aware that a particular message is a
reply to a previous one by checking the message identifiers in the set
of intentions that were suspended waiting for a reply. This feature
will be used when we give semantics to receiving $\TELL$ messages,
which can be sent by an agent when it spontaneously wants the receiver
to believe something (or at least to believe something about the
sender's beliefs), but can also be used when the agent receives an
``ask'' type of message and chooses to reply to it.


As mentioned earlier, it is not our aim to formalise every aspect of a
system of multiple AgentSpeak agents. We extend the previous semantics
only to the extent required to formalise speech-act based
communication for such agents.  It is relevant, therefore, to consider
a rule that defines message exchange as accomplished by the underlying
message exchange mechanism available in an overall agent architecture.
This is abstracted away in the semantics by means of the following
rule, where each $\AGCFGId{k}$, $k = 1 \ldots n$, is an agent
configuration $\CFGId{k}$:

\infrule[MsgExchg] {
 \msg{\MID,\ID_j,\ILF,\CNT} \in {M_{\ID_i}}_{Out}
}
{
 \{ \AGCFGId{1}, \ldots \AGCFGId{i}, \AGCFGId{j},
    \ldots \AGCFGId{n}, \ENV \} \longrightarrow \\
 \{ \AGCFGId{1}, \ldots \AGCFGpId{i}, \AGCFGpId{j},
    \ldots \AGCFGId{n}, \ENV \} \\[\rwsep]
\begin{array}{llcl}
\mbox{where: } & {M'_{\ID_i}}_{Out} & = & {M_{\ID_i}}_{Out} \setminus
                                       \{\msg{\MID,\ID_j,\ILF,\CNT} \} \\
               & {M'_{\ID_j}}_{In}  & = & {M_{\ID_j}}_{In} \cup
                                       \{\msg{\MID,\ID_i,\ILF,\CNT} \}
\end{array}
}

In the rule above, there are $n$ agents, and $\ENV$ denotes the
environment in which the agents are situated; typically, this is not
an AgentSpeak agent, it is simply represented as a set of properties
currently true in the environment and how they are changed by an
agent's actions. Note how, in a message that is to be sent, the second
component identifies the addressee (the agent to which the message is
being sent), whereas in a received message that same component
identifies the sender of the message.

\subsection{Speech-act Based Communication for AgentSpeak}

In this section we discuss the performatives that are most relevant
for communication in \as. These are largely inspired by corresponding
KQML performatives. We also consider some new performatives, related
to plan exchange rather than communication about propositions as
usual. The performatives that we consider are briefly described below,
where $s$ denotes the agent that sends the message, and $r$ denotes
the agent that receives the message. Note that \texttt{tell} and
\texttt{untell} can be used either for an agent to pro-actively send
information to another agent, or as replies to previous \texttt{ask}
messages.

\begin{description}
\item[\texttt{tell}:] $s$ intends $r$ to believe (that $s$ believes) the
  sentence in the message's content to be true;

\item[\texttt{untell}:] $s$ intends $r$ not to believe (that $s$ believes)
  the sentence in the message's content to be true;
  
\item[\texttt{achieve}:] $s$ requests that $r$ to intend to achieve a
  state of the world where the message content is true;

\item[\texttt{unachieve}:] $s$ requests $r$ to drop the
  intention of achieving a state of the world where the message
  content is true;
  
\item[\texttt{tell-how}:] $s$ informs $r$ of a plan (i.e., some
  know-how of $s$);
  
\item[\texttt{untell-how}:] $s$ requests $r$ to disregard a certain
  plan (i.e., to delete that plan from its plan library);
  
\item[\texttt{ask-if}:] $s$ wants to know if the content of the
  message is true for $r$;
  
\item[\texttt{ask-all}:] $s$ wants all of $r$'s answers to a question
  (i.e., all the beliefs that unify with the message content);
  
\item[\texttt{ask-how}:] $s$ wants all of $r$'s plans for a particular
  triggering event (in the message content).
\end{description}

For processing messages, a new selection function is necessary, which
operates in much the same way as the other selection functions
described in the previous section. The new selection function is
called $\SM$, and selects a message from $\MI$; intuitively, it
represents the priority assigned to each type of message by the
programmer. We also need another ``given'' function, but its purpose
is different from selection functions. The Boolean function
$\SOCACC(\ID,\ILF,\AT)$, where $\ILF$ is the illocutionary force of
the message from agent $\ID$, with propositional content $\AT$,
determines when a message is \emph{socially acceptable} in a given
context. For example, for a message of the form
$\langle\MID,\ID,\TELL,\AT\rangle$, the receiving agent may want to
consider whether $\ID$ is a relevant source of information, as even
remembering that $\ID$ believes $\AT$ might not be appropriate.  For a
message with illocutionary force $\ACHIEVE$, an agent would normally
check, for example, whether $\ID$ has sufficient social power over
itself, or whether it wishes to act altruistically towards $\ID$,
before actually committing to do whatever it is being asked.

We should mention that the role of $\SOCACC()$ in our framework is
analogous, on the receiver's side, to that of the ``cause to want''
and ``cause to believe'' predicates in Cohen and Perrault's plan-based
theory of speech acts \citeyear{cohen:79a}. That is, it provides a
bridge from the illocutionary force of a message to its perlocutionary
force.  The idea of having user-defined functions determining
relations such as ``trust'' and ``power'' has already been used in
practice by \citeauthor{BordiniMCAS} \citeyear{BordiniMCAS}.  Similar
interpretations for the use of $\SOCACC$ when applied to other types
of messages (e.g., \ASKIF) can easily be derived.

There is considerable work on elaborate conceptions of trust in the
context of multi-agent systems, for example in the work of
\citeauthor{CastPTMASCASIQ} \citeyear{CastPTMASCASIQ}.  In our
framework, more sophisticated notions of trust and power can be
implemented by considering the annotation of the sources of
information during the agent's practical reasoning rather than the
simple use of $\SOCACC$. The annotation construct facilitates
determining, in a plan context, the source of a belief before that
plan becomes an intended means.

Before we start the presentation of the semantic rules for
communication, it is worth noting that, in this paper in particular,
we do not consider \emph{nested} annotations. Nested annotations allow
the representation of beliefs about other agents' beliefs, or more
generally situations in which an agent $i$ was told $\varphi$ by $j$,
which in turn was told $\varphi$ by $k$, and so forth.

\subsection{Semantic Rules for Interpreting Received Messages}

\begin{paragraph*}{Receiving a Tell Message:}
  A $\TELL$ message might be sent to an agent either as a reply or as
  an ``inform'' action.  When receiving a $\TELL$ message as an inform
  (as opposed to a reply to a previous request), the \as agent will
  include the content of the received message in its knowledge base
  and will annotate the sender as a source for that belief. Note that
  this corresponds, in a way, to what is specified as the ``action
  completion'' condition by Labrou and Finin \citeyear{LabrouSAKQML}:
  the receiver will know about the sender's attitude regarding that
  belief. To account for the social aspects of multi-agent systems, we
  consider that social relations will regulate which messages the
  receiver will process or discard; this is referred to in the
  semantics by the \SOCACC\ function, which is assumed to be given by
  the agent designer. The rule shows that the annotated belief is
  added to the belief base, and the appropriate event is generated.

\infrule[Tell]
{
    \SM(\MI) = \msg{\MID,\ID,\TELL,\SB} \\
    (\MID,i) \not\in \MSI \mbox{ (for any intention $i$) } \\
    \SOCACC(\ID,\TELL,\SB)
}
{
    \CFG{\ProcMsg} \trans \CFGacmp{\SelEv} \\[\rwsep]
\begin{array}{llcl}
  \mbox{\emph{where:} }
                       & \MIli   & = & \MI \setminus \{\msg{\MID,\ID,\TELL,\SB}\} \\[\rwsep]
                       & \multicolumn{3}{l}{\mbox{and for each $b \in \SB$}:} \\
                       & \AGE'_{\BELS}  & = & \AGE_{\BELS}+b[\ID] \\
                       & \CEli   & = & \CE \cup
                       \{ \langle +b[\ID], \TR \rangle \}
\end{array}
}
\end{paragraph*}

\begin{paragraph*}{Receiving a Tell Message as a Reply:}
  This rule is similar to the one above, except that now the suspended
  intention associated with that particular message --- given that it
  is a reply to a previous ``ask'' message sent by this agent ---
  needs to be resumed. Recall that to resume an intention we just need
  to place it back in the set of intentions ($\CIli$).

\infrule[TellRepl]
{
    \SM(\MI) = \msg{\MID,\ID,\TELL,\SB} \\
    (\MID,i) \in \MSI \mbox{ (for some intention $i$) } \\
    \SOCACC(\ID,\TELL,\SB)
}
{
    \CFG{\ProcMsg} \trans \CFGacmp{\SelEv} \\[\rwsep]
\begin{array}{llcl}
  \mbox{\emph{where:} }
                       & \MIli   & = & \MI \setminus
            \{\msg{\MID,\ID,\TELL,\SB}\} \\
                       & \MSIli  & = & \MSI \setminus \{ (\MID,i) \} \\
                       & \CIli   & = & \CI \cup \{ i \} \\[\rwsep]
                       & \multicolumn{3}{l}{\mbox{and for each $b \in \SB$}:} \\
                       & \AGE'_{\BELS}  & = & \AGE_{\BELS}+b[\ID] \\
                       & \CEli   & = & \CE \cup
                       \{ \langle +b[\ID], \TR \rangle \}
\end{array}
}
\end{paragraph*}

\begin{paragraph*}{Receiving an Untell Message:}
  When receiving an $\UNTELL$ message, the sender of the message is
  removed from the set of sources giving accreditation to the atomic
  formula in the content of the message. In case the sender was the
  only source for that information, the belief itself is removed from
  the receiver's belief base. Note that, as the atomic formula in the
  content of an $\UNTELL$ message can have uninstantiated variables,
  each belief in the agent's belief base that can be unified with that
  formula needs to be considered in turn, and the appropriate events
  generated.

\infrule[Untell]
{
    \SM(\MI) = \msg{\MID,\ID,\UNTELL,\SAT} \\
    (\MID,i) \not\in \MSI \mbox{ (for some intention $i$) } \\
    \SOCACC(\ID,\UNTELL,\SAT)
}
{
    \CFG{\ProcMsg} \trans \CFGacmp{\SelEv} \\[\rwsep]
\begin{array}{llcll}
  \mbox{\emph{where:} } & \MIli   & = & \MI \setminus \{\msg{\MID,\ID,\UNTELL,\SAT}\}\quad \\[\rwsep]
                        & \multicolumn{3}{l}{\mbox{and for each }
                        b \in \{\AT\theta \mid} \\
                        & \multicolumn{3}{l}{\qquad \theta \in
                        \TEST(\AGBELS,\AT) \wedge
                        \AT \in \SAT\}}\\
                        & \AGE'_{\BELS}  & = & \AGE_{\BELS} - b[\ID] \\
                        & \CEli   & = & \CE \cup
                        \{ \langle -b[\ID], \TR \rangle \}
\end{array}
}
\end{paragraph*}

\begin{paragraph*}{Receiving an Untell Message as a Reply:}
  As above, the sender as source for the belief, or the belief itself,
  is excluded from the belief base of the receiver, except that now a
  suspended intention needs to be resumed (similarly to a $\TELL$ as a
  reply).

\infrule[UntellRepl] {
    \SM(\MI) = \msg{\MID,\ID,\UNTELL,\SAT} \\
    (\MID,i) \in \MSI \mbox{ (for some intention $i$) } \\
    \SOCACC(\ID,\UNTELL,\SAT)
}
{
    \CFG{\ProcMsg} \trans \CFGacmp{\SelEv} \\[\rwsep]
\begin{array}{llcll}
  \mbox{\emph{where:} } & \MIli   & = & \MI \setminus \{\msg{\MID,\ID,\UNTELL,\SAT}\}\quad \\
                        & \MSIli  & = & \MSI \setminus \{ (\MID,i) \} \\
                        & \CIli   & = & \CI \cup \{ i \} \\[\rwsep]
                        & \multicolumn{3}{l}{\mbox{and for each }
                        b \in \{\AT\theta \mid} \\
                        & \multicolumn{3}{l}{\qquad \theta \in
                        \TEST(\AGBELS,\AT) \wedge
                        \AT \in \SAT\}}\\
                        & \AGE'_{\BELS}  & = & \AGE_{\BELS}-b[\ID] \\
                        & \CEli   & = & \CE \cup
                        \{ \langle -b[\ID], \TR \rangle \}
\end{array}
}
\end{paragraph*}

\begin{paragraph*}{Receiving an Achieve Message:}
  In an appropriate social context (e.g., if the sender has ``power''
  over the receiver), the receiver will try to execute a plan whose
  triggering event is $+!\AT$; that is, it will try to achieve the
  goal associated with the propositional content of the message. An
  external event is thus included in the set of events (recall that
  external events have the triggering event associated with the empty
  intention $\TR$).

  Note that it is now possible to have a new focus of attention (a
  stack of plans in the set of intentions $I$) being initiated by the
  addition (or deletion, see below) of an achievement
  goal. Originally, only a belief change arising from perception of
  the environment initiated a new focus of attention; the plan chosen
  for that event could, in turn, have achievement goals in its body,
  thus pushing new plans onto the stack.

\infrule[Achieve]
{
    \SM(\MI) = \msg{\MID,\ID,\ACHIEVE,\AT} \\
    \SOCACC(\ID,\ACHIEVE,\AT)
}
{
    \CFG{\ProcMsg} \trans \CFGcmp{\SelEv} \\[\rwsep]
\begin{array}{llcl}
  \mbox{\emph{where:} } & \MIli   & = & \MI \setminus \{\msg{\MID,\ID,\ACHIEVE,\AT}\} \\
                        & \CEli   & = & \CE \cup
                          \{ \langle +!\AT, \TR \rangle \}
\end{array}
}

We shall later discuss in more detail the issue of autonomy. While
this gives the impression that simply accepting orders removes the
agent's autonomy (and similarly with regards to acquired beliefs), the
way the agent will behave once aware that another agent is attempting
to delegate a goal completely depends on the particular plans that
happen to be in the agent's plan library. If a suitable plan exists,
the agent could simply drop the goal, or could tell the interlocutor
that the goal delegation was noted but the goal could not be adopted
as expected, etc.

\end{paragraph*}

\begin{paragraph*}{Receiving an Unachieve Message:}
  This rule is similar to the preceding one, except that now the
  deletion (rather than addition) of an achievement goal is included
  in the set of events. The assumption here is that, if the agent has
  a plan with such a triggering event, then that plan should handle
  all aspects of dropping an intention.  However, doing so in practice
  may require the alteration of the set of intentions, thus requiring
  special mechanisms which have not been included in any formalisation
  of \as as yet, even though it is already available in practice, for
  example in the \Jason interpreter \cite{BordiniJASON}.

\infrule[Unachieve]
{
    \SM(\MI) = \msg{\MID,\ID,\UNACHIEVE,\AT} \\
    \SOCACC(\ID,\UNACHIEVE,\AT)
}
{
    \CFG{\ProcMsg} \trans \CFGcmp{\SelEv} \\[\rwsep]
\begin{array}{llcl}
  \mbox{\emph{where:} } & \MIli   & = & \MI \setminus \{\msg{\MID,\ID,\UNACHIEVE,\AT}\}\\
                        & \CEli   & = & \CE \cup \{ \langle -!\AT, \TR \rangle \}
\end{array}
}
\end{paragraph*}

\begin{paragraph*}{Receiving a Tell-How Message:}
  The \as notion of plan is related to Singh's concept of
  \emph{know-how} \citeyear{SinghMSTFIKHC}. Accordingly, we use the
  $\TELLHOW$ performative when agents wish to exchange know-how rather
  than communicate beliefs or delegate goals. That is, a $\TELLHOW$
  message is used by the sender (an agent or external source more
  generally) to inform an \as\ agent of a plan that can be used for
  handling certain types of events (as expressed in the plan's
  triggering event). If the source is trusted, the plans in the
  message content are simply added to the receiver's plan library.

\infrule[TellHow]
{
  \SM(\MI) = \msg{\MID,\ID,\TELLHOW,\SPL} \\
      (\MID,i) \not\in \MSI \mbox{ (for any intention $i$) } \\
  \SOCACC(\ID,\TELLHOW,\SPL)
}
{
    \CFG{\ProcMsg} \trans \CFGamp{\SelEv} \\[\rwsep]
\begin{array}{llcl}
  \mbox{\emph{where:} } & \MIli   & = & \MI \setminus \{\msg{\MID,\ID,\TELLHOW,\SPL}\} \\
                        & \AGE'_{\PLANS} & = & \AGE_{\PLANS} \cup \SPL
\end{array}
}

Note that we do not include any annotation to identify the source of a
plan, and so, with this semantics, it is not possible to take into
account the identity of the agent that provided a plan when deciding
whether to use it. In practice, this feature is implemented in the
\Jason interpreter, as the language is extended with the use of
annotated predicates as plan labels. This also allows programmers to
annotate plans with information that can be used for meta-level
reasoning (e.g., choosing which plan to use in case various applicable
plans are available, or which intention to execute next); examples of
such information would be the expected payoff of a specific plan and
its expected chance of success, thus allowing the use of
decision-theoretic techniques in making those choices.
\end{paragraph*}

\begin{paragraph*}{Receiving a Tell-How Message as a Reply:}
  The $\TELLHOW$ performative as a reply will also cause the suspended
  intention --- the one associated with the respective $\ASKHOW$
  message previously sent --- to be resumed.

\infrule[TellHowRepl]
{
    \SM(\MI) = \msg{\MID,\ID,\TELLHOW,\SPL} \\
    (\MID,i) \in \MSI \mbox{ (for some intention $i$) } \\
    \SOCACC(\ID,\TELLHOW,\SPL)
}
{
    \CFG{\ProcMsg} \trans \CFGacmp{\SelEv} \\[\rwsep]
\begin{array}{llcl}
  \mbox{\emph{where:} } & \MIli   & = & \MI \setminus \{\msg{\MID,\ID,\TELLHOW,\SPL}\} \\
                        & \MSIli  & = & \MSI \setminus \{ (\MID,i) \} \\
                        & \CIli   & = & \CI \cup \{ i \} \\
                         & \AGE'_{\PLANS} & = & \AGE_{\PLANS} \cup \SPL\\
\end{array}
}
\end{paragraph*}

\begin{paragraph*}{Receiving an Untell-How Message:}
  This is similar to the rule above, except that plans are now removed
  from the receiver's plan library. An external source may find that a
  plan is no longer appropriate for handling the events it was
  supposed to handle; it may then want to inform another agent about
  that. Thus, when receiving a socially acceptable $\UNTELLHOW$
  message, the agent removes the associated plans (i.e., those in the
  message content) from its plan library.

\infrule[UntellHow]
{
    \SM(\MI) = \msg{\MID,\ID,\UNTELLHOW,\SPL} \\
    \SOCACC(\ID,\UNTELLHOW,\SPL)
}
{
    \CFG{\ProcMsg} \trans \CFGamp{\SelEv} \\[\rwsep]
\begin{array}{llcl}
  \mbox{\emph{where:} } & \MIli   & = & \MI \setminus \{\msg{\MID,\ID,\UNTELLHOW,\SPL}\}\\
                        & \AGE'_{\PLANS} & = & \AGE_{\PLANS} \setminus \SPL
\end{array}
}
\end{paragraph*}

\begin{paragraph*}{Receiving an Ask-If Message:}
  The receiver will respond to this request for information if certain
  conditions imposed by the social settings (the \SOCACC\ function)
  hold between sender and receiver.
  
  Note that \emph{ask-if} and \emph{ask-all} differ in the kind of
  request made to the receiver. With the former, the receiver should
  just confirm whether the predicate in the message content is in its
  belief base or not; with the latter, the agent replies with all the
  predicates in its belief base that unify with the formula in the
  message content. The receiver processing an $\ASKIF$ message
  responds with the action of sending either a $\TELL$ (to reply
  positively) or $\UNTELL$ message (to reply negatively); the reply
  message has the same content as the $\ASKIF$ message. Note that a
  reply is only sent if the social context is such that the receiver
  wishes to consider the sender's request.

\infrule[AskIf]
{
    \SM(\MI) = \msg{\MID,\ID,\ASKIF,\{b\}} \\
    \SOCACC(\ID,\ASKIF,b)
}
{
    \CFG{\ProcMsg} \trans \CFGmp{\SelEv} \\[\rwsep]
\begin{array}{lcl}
     \mbox{\emph{where:} } \\
     \MIli & = & \MI \setminus \{\msg{\MID,\ID,\ASKIF,\{b\}}\}\\
     \MOli & = & \left\{
          \begin{array}{ll}
          \MO \cup \{ \msg{\MID,\ID,\TELL,\{b\}} \} & \mbox{if }
          \AGE_{\BELS} \models b \\
          \MO \cup \{ \msg{\MID,\ID,\UNTELL,\{b\}} \} \quad &
          \mbox{if } \AGE_{\BELS} \not\models b \\
          \end{array}
          \right.
\end{array}
}

The role that $\SM$ plays in the agent's reasoning cycle is slightly
more important here than originally conceived
\cite{MoreiraEOSBAOPLISABC-LNAI}. An agent considers whether to accept
a message or not, but the reply message is automatically assembled
when the agent selects (and accepts) any of the ``ask'' messages.
However, providing such a reply may require considerable computational
resources (e.g., the whole plan library may need to be scanned and a
considerable number of plans retrieved from it in order to produce a
reply message). Therefore, $\SM$ should normally be defined so that
the agent only selects an $\ASKIF$, $\ASKALL$, or $\ASKHOW$ message if
it determines the agent is not currently too busy to provide a reply.
\end{paragraph*}

\begin{paragraph*}{Receiving an AskAll:}
  As for $\ASKIF$, the receiver processing an $\ASKALL$ has to respond
  either with $\TELL$ or $\UNTELL$, provided the social context is
  such that the receiver will choose to respond.  As noted above, here
  the agent replies with all the predicates in the belief base that
  unify with the formula in the message content.

\infrule[AskAll]
{
    \SM(\MI) = \msg{\MID,\ID,\ASKALL,\{\AT\}} \\
    \SOCACC(\ID,\ASKALL,\AT)
}
{
    \CFG{\ProcMsg} \trans \CFGmp{\SelEv} \\[\rwsep]
}
$\begin{array}{lcl}
     \mbox{\emph{where:} } \\
      \MIli & = & \MI \setminus \{\msg{\MID,\ID,\ASKALL,\{\AT\}}\}\\
      \MOli & = & \left\{
        \begin{array}{ll}
        \MO \cup \{ \msg{\MID,\ID,\TELL,\SAT} \}, & \\
        \quad \SAT = \{ \AT\theta \mid \theta \in \TEST(\AGBELS,\AT) \} \quad
              & \mbox{if } \TEST(\AGBELS,\AT) \neq \{\} \\
        \MO \cup \{ \msg{\MID,\ID,\UNTELL,\{\AT\}} \} & \mbox{otherwise} \\
        \end{array}
        \right.
\end{array}$
\end{paragraph*}

\begin{paragraph*}{Receiving an AskHow:}
  The receiver of an $\ASKHOW$ has to respond with the action of
  sending a $\TELLHOW$ message, provided the social configuration is
  such that the receiver will consider the sender's request. In
  contrast to the use of $\UNTELL$ in $\ASKALL$, the response when the
  receiver knows no relevant plan (for the triggering event in the
  message content) is a reply with an empty set of plans.

\infrule[AskHow]
{
    \SM(\MI) = \msg{\MID,\ID,\ASKHOW,\TE} \\
    \SOCACC(\ID,\ASKHOW,\TE)
}
{
    \CFG{\ProcMsg} \trans \CFGmp{\SelEv} \\[\rwsep]
\begin{array}{lcl}
     \mbox{\emph{where:} } \\
     \MIli & = & \MI \setminus \{\msg{\MID,\ID,\ASKHOW,\TE}\}\\
     \MOli & = & \MO \cup \{ \msg{\MID,\ID,\TELLHOW,\SPL} \} \\
          & & \quad \mbox{and } \SPL = \{ p \mid (p,\theta) \in
     \RELPLANS(\AGPLANS,\TE) \}
\end{array}
}
\end{paragraph*}

\begin{paragraph*}{When $\SOCACC$ Fails:}
  All the rules above consider that the social relations between
  sender and receiver are favourable for the particular communicative
  act (i.e, they require $\SOCACC$ to be true). If the required social
  relation does not hold, the message is simply discarded --- it is
  removed from the set of messages and ignored. The rule below is used
  for receiving a message from an untrusted source, regardless of the
  performative.

\infrule[NotSocAcc]
{
    \SM(\MI) = \msg{\MID,\ID,\ILF,\SB} \\
    \neg \SOCACC(\ID,\ILF,\SB)\\
    (\mbox{with }\ILF \in \{\TELL,\UNTELL,\TELLHOW,\UNTELLHOW,\\
    \ACHIEVE,\UNACHIEVE,\ASKIF,\ASKALL,\ASKHOW\})
}
{
    \CFG{\ProcMsg} \trans \CFGmp{\SelEv} \\[\rwsep]
\begin{array}{llcl}
  \mbox{\emph{where:} }
                   & \MIli   & = & \MI \setminus \{\msg{\MID,\ID,\ILF,\SB}\}
\end{array}
}
\end{paragraph*}

\begin{paragraph*}{When $\MI$ is empty:}
  This last semantic rule states that, when the mail inbox is empty,
  the agent simply goes to the next step of the reasoning cycle
  (\SelEv).

\infrule[NoMsg] {
    \MI = \{\}
}
{
    \CFG{\ProcMsg} \trans \CFG{\SelEv} \\[\rwsep]
}
\end{paragraph*}

\subsection{Comments on Fault Detection and Recovery}
\label{atomfaultfailure}

As in any other distributed system, multi-agent systems can (and do)
fail in the real world. Possibly even more so than typical distributed
systems, given that multi-agent systems are normally used in dynamic,
unpredictable environments. In such contexts, failures are expected to
happen quite often, so agents need to recover from them in the best
possible way.  In the specific case of systems composed of \as agents,
failures can occur when, for instance, the agent for which a message
has been sent has left the multi-agent system, cannot be contacted, or
has ceased to exist (e.g., because of a machine or network crash).
Equally, an intention that was suspended waiting for a reply may never
be resumed again due to a failure in the agent that was supposed to
provide a reply. (However, note that \as agents typically have various
concurrent foci of attention --- i.e., multiple intentions currently
in the set of intentions --- so even if one particular intention can
never progress because another agent never replies, the agent will
simply carry on working on the other foci of attention.)

In the context of \as agents, both fault detection and recovery start
at the level of the infrastructure that supports the agent execution.
This infrastructure can adopt techniques available in traditional
distributed systems but with a fundamental difference: it is
responsible for adding appropriate events signaling failures in the
set $\CE$ of external events, or possibly resuming suspended
intentions and immediately making them fail if for example a message
reply has timed out. Such events, when treated by the agent in its
normal reasoning cycle, using the plan failure mechanism not
formalised here but available in practical interpreters, will trigger
a plan specifically written by the agent programmer which defines a
strategy for failure recovery. Therefore, from the point of view of
the formal semantics for \as, failure recovery reduces to event
handling and plan execution, and is partly the responsibility of the
underlying execution infrastructure and partly the responsibility of
programmers. We should note that various approaches for failure
detection and recovery within multi-agent systems in particular appear
in the literature \cite<e.g.,>{jennings:95a,KumarCohen00}. They
typically involve the use of special agents or plans defined to deal
with failure.

A natural concern when we have a set of agents executing concurrently
is that shared resources should always be left in a consistent state.
This is, of course, a classical problem in concurrency, and is
typically solved by atomically executing the parts of the code that
access the shared resources. Many programming language have constructs
that enable a programmer to guarantee the atomic execution of critical
sections. In a multi-agent system written in \as, atomicity is not
immediately an issue since there are no critical sections, given that
different \as agents do not directly share memory. However, \as agents
do exchange information in the form of messages, but the
responsibility for the management of such exchanges lies with the
underlying message passing infrastructure. On the other hand, agents
in a multi-agent systems typically share an environment, and if a
particular application requires environment resources to be shared by
agents, clearly programmers need to ensure that suitable agent
interaction protocols are used to avoid dead-/live-locks or
starvation.

Another possible source of concern regarding atomicity is the
concurrent execution of an agent's intentions. Agents can have several
intentions ready to be executed and each one of them can read/write
data that is shared with other intentions (as they all access the same
belief base). It is not in the scope of this paper to formalise
mechanisms to control such concurrency, but it is worth mentioning
that the \Jason interpreter provides programmers with the possibility
of annotating plans as being ``atomic'', so that when one of them is
selected for execution (see the \rname{SelInt} semantic rule), it is
guaranteed by the runtime agent platform that the plan execution will
not be suspended/interrupted (i.e., that no other intention will be
selected for execution in the following reasoning cycles) before the
whole plan finishes executing.

Next, we give an example intended to illustrate how the semantic rules
are applied during a reasoning cycle. The example includes agents
exchanging messages using the semantic framework for agent
communication formalised earlier in this section.


\section{Example of Reasoning Cycles of Communicating Agents}
\label{secERARC}

Consider the following scenario. Firefighting robots are in action
trying to control a rapidly spreading fire in a building, under the
supervision of a commander robot. Another robot is piloting a
helicopter to observe in which direction the fire is spreading most
rapidly.  Let the robot in the helicopter be \ra, let \rb be the
ground commander, and let \rc be one of the firefighting robots.

One of the plans in \ra's plan library, which we shall refer to as
\psa, is as shown in Figure~\ref{fig:ffrps}. This plan says that as
soon as \ra perceives fire spreading in direction \texttt{D}, it tells
\texttt{R} that fire is spreading towards \texttt{D}, where \texttt{R}
is the agent it believes to be the ground commander. Plan \psb is one
of the plans that robot \rb (the commander) has in its plan library,
which is also shown in Figure~\ref{fig:ffrps}. Plan \psb says that,
when \rb gets to believe\footnote{Note that, because the plan's
  triggering event does not require a particular source for that
  information (as in, e.g., \texttt{+spreading(D)[percept]}), this
  plan can be used when such belief is acquired from communication as
  well as perception of the environment.} that fire is spreading in
direction \texttt{D}, it will request the robot believed to be closest
to that part of the building to achieve a state of the world in which
\texttt{D} is the fighting post of that robot (in other words, that
the robot should relocate to the part of the building in direction
\texttt{D}).

\begin{figure}[ht]
\begin{center}
\fbox{\parbox{\linewidth}{
\ttfamily
\begin{tabbing}
+te\=: \=<- \=\kill
\underline{\textrm{\textbf{\ra's plan \psa}}}\\
\\
+spreading(D) \\
\> :  commander(R); \\
\> \> <- .send(R,tell,spreading(D)).\\
\\
\underline{\textrm{\textbf{\rb's plan \psb }}}\\
\\
+spreading(D) \\
\> :  closest(D,A); \\
\> \> <-.send(A,achieve,fight\_post(A,D)).
\end{tabbing}}}
\end{center}
\caption{Plans used in the firefighting robots example.}
\label{fig:ffrps}
\end{figure}

We now proceed to show how the rules of the operational semantics
apply, by using one reasoning cycle of the \as agent that controls \ra
as an example; the rules for communication will be exemplified
afterwards.  For simplicity, we assume \ra is currently defined by a
configuration $\langle\AGEA,\CCA,\MA,\TA,\sA\rangle$, with $\sA =
\ProcMsg$ and the $\AGEA$ component having:

\[ \AGEAbels = \{ \atom{commander(\rb)} \}\mbox{, and} \]
\[ \AGEAplans = \{ \psa \}. \]

Suppose \ra has just perceived fire spreading towards the south. After
the belief revision function (see Section~\ref{ssecSe}) has operated,
\ra's beliefs will be updated to $\AGEAbels = \{
\atom{commander(\rb)}, \atom{spreading(south)} \}$ and the $\CAE$
component of \ra's configuration (i.e., its set of events) will be as
follows:

\[ \CAE = \{ \langle\atom{+spreading(south)},\TR\rangle \}. \]

At this point, we can show the sequence of rules that will be applied
to complete one reasoning cycle: see Table~\ref{tab:SeqRulesRC}, where
the left column shows the rule being applied and the right column
shows \emph{only the components of the configuration which have
  changed} as a consequence of that rule having been applied. Note
that the ``next step'' component $\sA$ changes in an obvious way
(given the rules being applied), so we only show its change for the
very first step, when there are no messages to be processed and the
cycle goes straight on to selecting an event to be handled in that
reasoning cycle.

\begin{table}[ht]
\caption{Example sequence of rules applied in one reasoning cycle.}
\label{tab:SeqRulesRC}
\begin{center}
$\begin{array}{|c@{~}|@{~~}l|}
\hline
\mbox{\textbf{Rule}} & \mbox{\textbf{Changed Configuration Components}} \\
\hline
\hline
\rname{NoMsg}        & \sA = \SelEv \\
\hline
\rname{SelEv$_{1}$}  & \CAE = \{\} \\
                     & \TAeps = \langle\atom{+spreading(south)},\TR\rangle \\
\hline
\rname{Rel$_{1}$}    & \TAr = \{ (\psa,\theta_R) \}\mbox{, where }
       \theta_R=\{\atom{D}\mapsto\atom{south}\} \\
\hline
\rname{Appl$_{1}$}   & \TAap = \{ (\psa,\theta_A) \}\mbox{, where }
       \theta_A=\{\atom{D}\mapsto\atom{south}, \atom{R}\mapsto\rb\} \\
\hline
\rname{SelApp}       & \TArho = (\psa,\theta_A) \\
\hline
\rname{ExtEv}        & \CAI = \{ [\psa\theta_A] \} \\
\hline
\rname{SelInt$_{1}$} & \TAiota = [\psa\theta_A] \\
\hline
~\rname{ExecActSnd}   & \MAout =
       \{ \msg{\MID_1,\rb,\TELL,\{\atom{spreading(south)}}\} \} \\
                       & \CAI =
       \{ [\atom{+spreading(south) : commander(\rb) <- \TR}]\} \\
\hline
\rname{ClrInt$_{1}$} & \CAI = \{\}\\
\hline
\end{array}$
\bigskip
\end{center}
\end{table}

After \ra's reasoning cycle shown in the table, rule \rname{MsgExchg}
applies and, assuming $\langle\AGEB,\CCB,\MB,\TB,\sB\rangle$ is \rb's
configuration, which for simplicity we assume is the initial (i.e.,
empty) configuration hence $\sB=\ProcMsg$, we shall have that:

\[ \MBin = \{\msg{\MID_1,\ra,\TELL,\{\atom{spreading(south)}\}}\}, \]

\noindent
which then leads to rule \rname{Tell} being applied, thus starting a
reasoning cycle (similar to the one in Table~\ref{tab:SeqRulesRC}) in
\rb from a configuration that will have had the following components
changed (see Rule~\rname{Tell}):

\[ \MBin = \{\} \]
\[ \AGEBbels = \{ \atom{spreading(south)[{\ra}]} \} \]
\[ \CBE = \{ \langle\atom{+spreading(south)[{\ra}]},\TR\rangle \}. \]

After a reasoning cycle in \rb, we would have for \rc that:

\[ \begin{array}{rcl}
\MCin & = & \{
\msg{\MID_2,\rb,\ACHIEVE,\atom{fight\_post(\rc,south)}} \},
    \end{array} \]

\noindent
where $\MCin$ is \rc's mail inbox (and assuming $\MCin$ was previously
empty). Note that \rc's $\SOCACC$ function used by rule
\rname{Achieve} (leading to $\MID_2$ being included in $\MCin$ as
stated above) would probably consider the hierarchy determined by the
firefighters' ranks. Robot \rc would then consider the events
generated by this received message in its subsequent reasoning cycles,
and would act in accordance to the plans in its plan library, which we
do not show here, for simplicity.

\section{Developing More Elaborate Communication Structures}
\label{secDEFC}

We sometimes require more elaborate communication structures than
performatives such as those discussed in Section~\ref{secSCAA}. On the
other hand, it is of course important to keep any communication scheme
and its semantic basis as simple as possible. We emphasise that, in
our approach, more sophisticated communication structures can be
programmed, on top of the basic communication features formalised
here, through the use of plans that implement interaction protocols.
In practical \as interpreters, such communication features (built by
composing of the atomic performatives) can be provided to programmers
either as extra pre-defined performatives or as plan templates in plan
libraries made publicly available --- in \Jason \cite{BordiniJASON},
the former approach has been used, but the latter is also possible. In
this section, we give, as examples of more advanced communication
features, plans that allow agents to reach shared beliefs and ensure
that agents are kept informed of the adoption of their goals by other
agents. Note however that the examples make use of a simple practical
feature (available, e.g., in \Jason) which does not appear in the
abstract syntax we used earlier in the formal presentation: a variable
instantiated with a first order term can be used, within certain
constructs (such as belief or goal additions), in place of an atomic
formula, as usual also in Prolog implementations.

\begin{exmp}[Shared Beliefs]\label{exSB}
  
  If a network infrastructure is reliable, it is easy to ensure that
  agents reach shared beliefs. By reaching a shared belief, we mean
  two agents believing $b$ as well as believing that the other agent
  also believes $b$. More explicitly, we can say agents \texttt{ag1}
  and \texttt{ag2} share belief $b$ if \texttt{ag1} believes both
  $b[$\texttt{self}$]$ and $b[$\texttt{ag2}$]$, at the same time that
  \texttt{ag2} believes both $b[$\texttt{self}$]$ and
  $b[$\texttt{ag1}$]$.  In order to allow agents \texttt{ag1} and
  \texttt{ag2} to reach such shared beliefs, it suffices\footnote{Even
    if both agents do not have such plans in advance, but are willing
    to be told how to reach shared beliefs (by accepting $\TELLHOW$
    messages from agents who have such know-how), they can become
    capable of reaching shared beliefs too.} to provide both agents
  with copies of the following plans:

\begin{alltt}
\underline{\textbf{rsb1}}
+!reachSharedBel(P,A) : not P[self]
  <- +P;
     !reachSharedBel(P,A).

\underline{\textbf{rsb2}}
+!reachSharedBel(P,A) : P[self] & not P[A]
  <- .send(A,tell,P);
     .send(A,achieve,reachSharedBel(P,\(me\))).

\underline{\textbf{rsb3}}
+!reachSharedBel(P,A) : P[self] & P[A]
  <- true.
\end{alltt}
\end{exmp}

In the plans above, $me$ stands for the agent's own name. (Recall
that, as in Prolog, an uppercase initial denotes a logical variable.)
Assume agent \texttt{ag1} has the above plans and some other plan, an
instance of which is currently in its set of intentions, which
requires itself and \texttt{ag2} to share belief \texttt{p(X)}. Such a
plan would have the following goal in its body:
\texttt{!reachSharedBel(p(X),ag2)}. This would eventually lead to the
execution of the plans in the example above, which can now be
explained. The plan labelled \texttt{rsb1} says that if \texttt{ag1}
has a (new) goal of reaching a shared belief \texttt{P} with agent
\texttt{A}, in case \texttt{ag1} does not yet believe \texttt{P}
itself, it should first make sure itself believes \texttt{P} --- note
that `\texttt{+P;}' in the body of that plan will add the ground
predicate bound to \texttt{P} with source \texttt{self} as a new
belief to agent \texttt{ag1} --- then it should again have the goal of
reaching such shared belief (note that this is a recursive plan). This
time, plan \texttt{rsb1} will no longer be applicable, so
\texttt{rsb2} will be chosen for execution. Plan \texttt{rsb2} says
that, provided \texttt{ag1} believes \texttt{P} but does not yet
believe that agent \texttt{A} believes \texttt{P}, it should tell
agent \texttt{A} that itself (\texttt{ag1}) believes \texttt{P}, then
finally ask \texttt{A} to also achieve such shared belief with
\texttt{ag1}.

Agent \texttt{ag2}, which also has copies of the plans in the example
above, would then, given the appropriate $\SOCACC$ function, have an
instance of plan \texttt{rsb1} in its own set of intentions, and will
eventually execute \texttt{rsb2} as well, or directly \texttt{rsb2} as
the case may be. Note that the last line of plan \texttt{rsb2}, when
executed by the agent that was asked to reach a shared believe, rather
than the one who took the initiative, is redundant and will lead the
other agent to using \texttt{rsb3}, which only says that no further
action is required, given that the shared belief has already been
obtained. Clearly, there are more efficient ways of implementing a
protocol for reaching shared belief, but we present this because the
same plans can be used regardless of whether the agent takes the
initiative to reach a shared belief or not. The version we give here
is therefore arguably more elegant, and its symmetry facilitates
reasoning about the protocol.

We now give another example, which shows how agents can have further
information about requests for goal adoption (i.e., when they ask
another agent to achieve some state of affairs on their behalf).

\begin{exmp}[Feedback on Goal Adoption]
\label{exFGA}

It is often the case that, if one agent asks another agent to do
something, it may want to have at least some reassurance from the
other agent that it has agreed to do whatever it has been asked.
Furthermore, it may want to know when the other agent believes it has
accomplished the task. The following plans can be used for
\texttt{ag1} to delegate tasks to \texttt{ag2} in such a way.

\newcommand{\ctone}{\ensuremath{cntxt_1}}
\newcommand{\cttwo}{\ensuremath{cntxt_2}}
\begin{alltt}
\textrm{\textbf{ag1 plans:}}

\underline{\textbf{nsd1}}
+needDoneBy(G,A) : not delegatedTo(G,A)
  <- +delegatedTo(G,A);
     .send(A,achieve,doAndFeedbackTo(G,\(me\))).

\underline{\textbf{nsd2}}
+needDoneBy(G,A)
  : agreedToDo(G)[A] & not finishedDoing(G)[A]
    <- .send(A,tell,shouldHaveFinished(G));
       ...

\underline{\textbf{nsd3}}
+needDoneBy(G,A) : finishedDoing(G)[A]
  <- ...

...

\underline{\textbf{fd}}
+finishedDoing(G)[A] : true
  <- -delegatedTo(G,A);
     -agreedToDo(G)[A].


\textrm{\textbf{ag2 plans:}}

\underline{\textbf{dft1}}
+!doAndFeedbackTo(G,A) : \ctone
  <- .send(A,tell,agreedToDo(G));
     +!G;
     .send(A,tell,finishedDoing(G)).

\underline{\textbf{dft2}}
+!doAndFeedbackTo(G,A) : \cttwo
  <- .send(A,tell,cannotDo(G)).
\end{alltt}
\end{exmp}

In the example above, we assume that something perceived in the
environment leads the agent to believe that it needs some goal
\texttt{G} to be achieved by agent \texttt{A}, and that such
perception recurs at certain intervals, when the need that motivated
the request still exists and the result of \texttt{A}'s achieving
\texttt{G} has not been observed. Plan \texttt{nsd1} is used when such
a need occurs but no request has been as yet sent to \texttt{A}. The
plan ensures that \texttt{ag1} will remember that it already asked
\texttt{A} (say, \texttt{ag2}) to do \texttt{G} and then that agent to
achieve a goal associated with a special plan: see plan \texttt{dft1}
in \texttt{ag2}. Such plan makes sure that the requesting agent is
informed both that \texttt{ag2} has adopted the goal as requested
(before it attempts to achieve it) as well as when the agent believes
to have achieved \texttt{G}. The programmer should define the
$\SOCACC$ function so that \texttt{ag2} accepts such requests from
\texttt{ag1}, but the programmer can still determine how autonomous
\texttt{ag2} will be by using appropriate plan contexts. In plan
\texttt{dft1}, context $cntxt_1$ would determine the circumstances
under which agent \texttt{ag2} believes it will be able to adopt the
goal, and context $cntxt_2$, in plan \texttt{dft2}, can be used for
the circumstances in which \texttt{ag2} should simply inform it will
not adopt the goal as requested by \texttt{ag1} (a more elaborate plan
could explain why the agent cannot adopt the goal, for example in case
there are more than one situation in which the goal cannot be
adopted).

Going back to plans \texttt{nsd2} and \texttt{nsd3} in agent
\texttt{ag1}, the former is used to ``put pressure'' on the agent that
has adopted \texttt{ag1}'s goal \texttt{G}, as the need for that has
been perceived again and \texttt{A} has already agreed to do that, so
presumably it is not doing it fast enough. Clearly, the
``\texttt{shouldHaveFinished}'' belief should trigger some plan in
\texttt{ag2} for it to have the desired effect. Plan \texttt{nsd3} is
just a template for one of various alternative courses of actions to
be taken by \texttt{ag1} when the need that motivated a request for
\texttt{ag2} to adopt a goal still exists but \texttt{ag2} believes
the goal has already been achieved: that might be an old belief which
needs to be revised and a new request made, or \texttt{ag1} could try
asking another agent, or inform \texttt{ag2} that its belief about
achieving \texttt{G} might be wrong, etc. Plan \texttt{fd} is used
simply to remove unnecessary beliefs used in previous stages of the
interaction aimed at a goal adoption.

It is not difficult to see that plans for other important multi-agent
issues, such as ensuring agents are jointly committed to some course
of action, can be developed by elaborating on the combinations of
communication performatives along the lines of the examples above. On
the other hand, many other complications related to agent interaction
might need to be accounted for which could not be addressed in the
simple examples provided here, such as shared beliefs becoming
inaccurate with the passage of time. Further plans to go with the ones
shown here would be required for coping with such complications, when
necessary in particular applications.

\section{Proving Communication Properties of \as Agents}
\label{secPCPAA}

\citeauthor{BordiniPBPAOPL-ATPA} \citeyear{BordiniPBPAOPL-ATPA}
introduced a framework for proving BDI properties of \as agents based
on its operational semantics. The framework included precise
definitions of how the BDI modalities are interpreted in terms of
configurations of the transition system that gives semantics to \as.
Those same definitions are used in the work on model checking for \as
\cite{BordiniMCRA}, which allows the use of automated techniques for
verification of \as programs. Below, we give an example of a proof
using the operational semantics for a simple property that involves
only the \emph{belief} modality. As the belief modality is very clear
with respect to an \as agent, given that its architecture includes a
belief base explicitly, we avoid the need to discuss in this paper our
previous work on the interpretation of the modalities
\cite{BordiniPBPAOPL-ATPA}.

\begin{prop}[Reachability of Shared Beliefs]
  If any two \as agents $ag_1$ and $ag_2$ have in their plan libraries
  the \texttt{rsb1}, \texttt{rsb2}, and \texttt{rsb3} plans shown in
  Example~\ref{exSB}, and they also have an appropriate $\SOCACC$
  function as well as the usual implementation of selection functions
  (or others for which fairness is also guaranteed, in the sense that
  all events and intentions are eventually selected), if at some
  moment in time $ag_1$ has \texttt{reachSharedBel($b$,$ag_2$)} as a
  goal in its set of events (i.e., it has an event such as
  $\langle$\texttt{+!reachSharedBel($b$,$ag_2$)}$, i \rangle$, with
  $i$ an intention), then \emph{eventually both agents will believe
    $b$ and believe that the other agent also believes $b$} --- note
  that this can be formulated using a BDI-like logic on top of LTL as
  \[ \evently (\Bel{ag_1}{b[self]} \wedge \Bel{ag_2}{b[ag_1]} \wedge
  \Bel{ag_2}{b[self]} \wedge \Bel{ag_1}{b[ag_2]}). \]
\end{prop}

\begin{pf}
  It is assumed that $ag_1$ has $\langle
  \mathtt{+!reachSharedBel}(b,ag_2), i \rangle$ in its set of events.
  Assume further that this is precisely the event selected when rule
  \rname{SelEv$_{1}$} is applied. Then rule \rname{Rel$_{1}$} would
  select plans \texttt{rsb1}, \texttt{rsb2}, and \texttt{rsb3} as
  relevant for the chosen event. Rule \rname{Appl$_{1}$} would narrow
  this down to \texttt{rsb1} only as, presumably, $ag_1$ does not yet
  believe $b$ itself. Rule \rname{SelAppl} would necessarily select
  \texttt{rsb1} as intended means, given that it is the only
  applicable plan, and rule \rname{IntEv} would include
  $i[\mathtt{rsb1}]$ in the set of intentions (i.e., the chosen
  intended means would be pushed on top of the intention that
  generated the event above). Consider now that in this same reasoning
  cycle (for simplicity), rule \rname{SelInt$_{1}$} would choose
  precisely that intention for execution within this reasoning cycle.
  Then rule \rname{AddBel} would add $b[self]$ to $ag_1$'s belief
  base, hence $\Bel{ag_1}{b[self]}$.
  
  In subsequent reasoning cycles, when $ag_1$'s intention selection
  function selects the above intention for further execution, rule
  \rname{AchvGl} would generate again an internal event $\langle
  \mathtt{+!reachSharedBel}(b,ag_2), i \rangle$. The process is then
  as above expect that plan \texttt{rsb1} is no longer applicable, but
  \texttt{rsb2} is, and is therefore chosen as intended means.  When
  that plan is executed (similarly as described above), rule
  \rname{ExecActSnd} would add message $\langle mid_{1}, ag_{2},
  \TELL, b\rangle$ to $ag_1$'s $\MO$ component. Rule \rname{MsgExchg}
  then ensures that message $\langle mid, ag_{1}, \TELL, b\rangle$ is
  added to $ag_2$'s $\MI$ component, which in the beginning of the
  next reasoning cycle would lead to rule \rname{Tell} adding
  $b[ag_1]$ to $ag_2$'s belief base, hence $\Bel{ag_2}{b[ag_1]}$. When
  the intention is selected for execution in a third reasoning cycle,
  the final formula in the body of plan \texttt{rsb1} would be
  executed. By the use of similar rules for sending and receiving
  messages, we would have $ag_2$ receiving a message $\langle mid_{2},
  ag_1, \ACHIEVE, \mathtt{reachSharedBel}(b,ag_1)\rangle$, so now rule
  \rname{Achieve} is used for interpreting the illocutionary force in
  that message, thus adding an event $\langle
  \mathtt{+!reachSharedBel}(b,ag_1), i \rangle$ to $ag_2$'s set of
  events. Note that this is precisely how the process started in
  $ag_1$ so the same sequence of rules will apply to $ag_2$, which
  will, symmetrically, lead to $\Bel{ag_2}{b[self]}$ and
  $\Bel{ag_1}{b[ag_2]}$ being true, eventually. At that point in time
  we will have $(\Bel{ag_1}{b[self]} \wedge \Bel{ag_2}{b[ag_1]} \wedge
  \Bel{ag_2}{b[self]} \wedge \Bel{ag_1}{b[ag_2]})$.
\end{pf}
  
As discussed earlier, because $ag_2$ is using exact copies of the
plans used by $ag_1$, $ag_2$ will also ask $ag_1$ to reach $b$ as a
shared belief, even though $ag_1$ has already executed its part of the
joint plan. This is why plan \texttt{rsb3} is important. It ensures
that the agent will act no further when its own part of the joint plan
for reaching a shared belief has already been achieved.

Note, however, that it is only possible to guarantee that a shared
belief is reached in \emph{all possible runs} if neither agent has
plans that can interfere negatively with the execution of the plans
given in Example~\ref{exSB}, for example by forcing the deletion of
any instance of belief $b$ before such shared belief is reached. This
is a verification exercise different from the proposition we wanted to
prove, showing that shared beliefs \emph{can} be reached (under the
given assumptions).

\section{Applications of \as and Ongoing Work}
\label{secApp}

We mention here some of the applications written in \as. The \as
programming language has also been used in academia for student
projects in various courses. It should be noted, however, that the
language is clearly suited to a large range of applications for which
it is known that BDI systems are appropriate; various applications of
PRS \cite{GeorRRP} and dMARS \cite{KinnyDMARSALS}, for example, have
appeared in the literature \cite[Chapter~11]{WoolIMAS}.

One particular area of application in which we have great interest is
Social Simulation \cite{DoranSSI}. In fact, \as is being used as part
of a project to produce a platform tailored particularly for social
simulation. The platform is called MAS-SOC is being developed by
\citeauthor{BordiniMAS-SOCsspaop} \citeyear{BordiniMAS-SOCsspaop}; it
includes a high-level language called ELMS
\cite{OkuyamaELMSedlmas-LNAI} for describing environments to be shared
by multiple agents. This approach was used to develop, for example, a
social simulation on social aspects of urban growth \cite{KraftaCOHA}.
Another area of application that has been initially explored is the
use of \as for defining the behaviour of animated characters for
computer animation or virtual reality environments
\cite{TorresAAMFAVE}.

More recently, \as has been used in the implementation of a team of
``gold miners'' as an entry to an agent programming competition
\cite{BordiniUJITGMpd}. In this scenario\footnote{See
  \url{http://cig.in.tu-clausthal.de/CLIMAContest/} for details.},
teams of agents must coordinate their actions in order to collect as
much gold as they can and to deliver the gold to a trading agent
located in a depot where the gold is safely stored. The \as team,
composed of four mining agents and one leader that helped coordinate
the team of miners, won the competition in 2006. It is worth noting
that the language support for high-level communication (formalised in
this paper) proved to be an important feature for designing and
implementing the system.

The \as interpreter and multi-agent platform \Jason is being
constantly improved, with the long term goal of supporting various
multi-agent systems technologies.  An important aspect of \Jason is
precisely that of having formal semantics for most of its essential
features.  Various projects are currently looking at extending \Jason
in various ways, for example to combine it with an organisational
model such as the one propose by \citeauthor{HubnerUMCFMR}
\citeyear{HubnerUMCFMR}. This is particularly important given that
social structure is a fundamental notion for developing complex
multi-agent systems. Another area of development is to incorporate
ontologies into an \as belief base
\cite{MoreiraAOPUOR-LNCS,VieiraAOPLCC}, facilitating the use of \Jason
for Semantic Web applications. Recent work has also considered
automated belief revision \cite{AlechinaABRAS} and plan exchange
mechanisms \cite{AnconaCooAScape}. A more detailed description of the
language and comparison with other agent-oriented programming
languages was given by \citeauthor{BordiniJGFAOP}
\citeyear{BordiniJGFAOP}.

\section{Conclusions}
\label{secC}

As pointed out by \citeauthor{SinghACLRP} \citeyear{SinghACLRP}, there
are various perspectives for the semantics of agent communication.
Whereas the sender's perspective is the most common one in the
literature, our approach uses primarily that of the receiver.  We have
given a formal semantics to the processing of speech-act based
messages by an \as agent. Previous attempts to define the semantics of
agent communication languages \cite<e.g.,>{LabrouSAKQML} were based on
the ``pre-condition -- action -- post-condition'' approach, referring
to agent mental states in modal languages typically based on Cohen and
Levesque's work on intention \citeyear{CohenICC}. Our semantics for
communication, besides being more closely linked to implementation (as
it serves as the specification for an interpreter for an agent
programming language), can also be used in the proof of communication
properties \cite{WoolSIVACL}.

Our work is somewhat related to that of \citeauthor{BoerFSEIMAS}
\citeyear{BoerFSEIMAS} and \citeauthor{EijkVFAC} \citeyear{EijkVFAC},
which also provide an operational semantics for an agent communication
language.  However, their work does not consider the effects of
communication in terms of BDI-like agents (such as those written in
\as). The idea of giving semantics to speech-act based communication
within a BDI programming language was first introduced by
\citeauthor{MoreiraEOSBAOPLISABC-LNAI}
\citeyear{MoreiraEOSBAOPLISABC-LNAI}. Subsequently,
\citeauthor{DastaniCGDA} \citeyear{DastaniCGDA} also published some
initial work on the semantics of communication for~3APL agents,
although with the emphasis being on formalising the message exchange
mechanisms for synchronous and asynchronous communication. In
contrast, we largely abstract away from the specific message exchange
mechanism (this is formalised at a very high level in our semantics),
and we are interested only in asynchronous communication (which is the
usual communication model for cognitive agents). In order to
illustrate their message exchange mechanism, Dastani et al.\ gave
semantics to the effects of receiving and treating ``request'' and
``inform'' messages --- that is, they only consider information
exchange. Our work uses a much more comprehensive selection of
illocutionary forces, and the main contribution is precisely in giving
detailed semantics to the ways in which the various illocutionary
forces affect the mental states of agents implemented in a programming
language which actually has precise definitions for the notions of the
BDI architecture.  A denotational semantics for agent communication
languages was proposed by \citeauthor{Guerin01} \citeyear{Guerin01},
but the semantics is given for an abstract version of an ACL and does
not address the issues of interaction between an ACL and other
components of an agent architecture.

In this paper we provided new semantic rules for all the illocutionary
forces used in a communication language for \as agents. In giving
semantics to communicating \as\ agents, we have provided the means for
the implementation of \as\ interpreters with such functionality, as
well as given a more computationally grounded semantics of speech-act
based agent communication. In fact, the operational semantics
presented in this paper proved useful in the implementation of \as\ 
interpreters such as \Jason \cite{BordiniJASON}. While Singh's
proposal for a social-agency based semantics \citeyear{SinghACLRP} may
be appropriate for general purpose agent communication languages such
as FIPA or KQML, within the context of a BDI agent programming
language, our approach can be used without any of the drawbacks
pointed out by Singh.

The fact that we have to deal with the intentional states of other
agents when giving semantics of communication leads us to a number of
related pragmatic questions. First, many treatments of speech-act
style communication make use of \emph{mutual} mental states --- mutual
belief, common knowledge, and similar. We do not make use of mutual
mental states in our formalisation.  There are good reasons for this.
First, although mutual mental states are a useful and elegant tool for
\emph{analysis}, it is known that they represent \emph{theoretical
  idealisations} only, which \emph{cannot be achieved in systems which
  admit the possibility of message delivery failure}
\cite{halpern:90a}. Thus, although mutual mental states are a useful
abstraction for understanding how communication works, they cannot,
realistically, be implemented, as there will always be a mismatch
between the implementation (which excludes the possibility of mutual
mental states being faithfully implemented) and the theory. This is
primarily why mutual mental states form no part of our language or
semantics, but are built on top of the fundamental communication
primitives that we formalised in this paper, as shown in
Section~\ref{secDEFC}. Note that it is also known that mutual mental
states can be \emph{simulated}, to any desired degree of nesting, by
an appropriate message acknowledgement scheme \cite{halpern:92b},
therefore in our approach this problem can be solved by mechanisms
such as processed messages triggering the action of sending a message
that acknowledges receipt. It is also worth adding that the belief
annotation scheme used in our language permits agents to have a simple
mechanism for nested beliefs: the annotation of source in a belief is
an indication that the agent who sent the message believed in its
propositional content at the time the message was sent (but note that
this is an indication only, unless agent veracity is guaranteed).
Annotation of information source at the time a message is received is
done automatically according to the semantics we have given. However,
programmers can also use the belief base to register sent messages,
possibly using annotations in the same manner as for received
messages.  These would function as an indication of other agents'
states of mind, but from the point of view of the sender. We plan to
deal with these questions which lie in the gray area between semantics
and pragmatics in more detail in future work.

While discussing models of mutual mental states, we should also
mention in passing that \emph{joint intentions} do not form part of
our semantics, although they are widely used in the implementation of
coordination schemes for multi-agent systems, following the seminal
work of \citeauthor{levesque:90a} \citeyear{levesque:90a}. The fact
that such constructs are not built into the language (or the language
semantics) as primitives does not preclude them being
\emph{implemented} using the language constructs, provided the usual
practical considerations and assumptions, such as limiting the number
of required acknowledgement messages for the achievement of shared
beliefs, are in place. Indeed, this is exactly the approach taken by
Tambe, in his STEAM system \citeyear{tambe:97a}, and Jennings, in his
GRATE* system \citeyear{jennings:95a}. The examples in
Section~\ref{secDEFC} help indicate how this can be achieved by
further elaboration of those plans, making use of the communication
primitives for which we gave semantics in this paper.

We anticipate that readers will ponder whether our semantics limits
the autonomy of agents that use our approach to communication.  We
provide the $\SOCACC()$ function which works as an initial ``filter'',
but this may give the impression that beliefs are just
acquired/trusted and goals adopted after such simple filter. It is
very important to emphasise that the actual \emph{behaviour} of the
agent ensuing from communication received from other agents completely
depends on the particular plans the agent happens to have in its plan
library; in the current semantics, only the ``ask'' variants,
$\TELLHOW$, and $\UNTELLHOW$ performatives are dependent solely on the
$\SOCACC$ filter. In Example~\ref{exFGA}, we mentioned that some plan
contexts should be used to determine whether the agent would actually
act towards achieving a goal as requested by another agent, or choose
not to commit to achieving the goal. This is the general rule: the
agent autonomy depends on the plans given by the agent programmer or
obtained by communication with other agents (the plans currently in
the agent's plan library). It would be typically the programmer's
responsibility to write plans that ensure that an agent will be
``sufficiently autonomous'' for its purpose in a given application or,
to use a more interesting notion, to program agents with
\emph{adjustable autonomy}. Similarly, how benevolent or
self-interested an agent will be, and to what extent beliefs acquired
from other agents are to be trusted, are all issues that
\emph{programmers} have to be careful about: the semantics of
communication itself does not ensure one case or the other. Needless
to say, it will be a much more difficult task to program agents to
take part in open systems where other agents are self-interested and
cannot be trusted. While an agent programming language combined with a
suitable agent communication language gives much support for such
task, it surely does not automatically solve all those problems; it
still remains a complex programming task.

It is also worth commenting on how our semantics can be used by other
researchers, particularly those using agent programming languages
other than AgentSpeak. The main point here is that our semantics
provides a \emph{reference} to the semantics of the communication
language used in the context of agent-oriented programming. That is,
using our semantics, it is possible to predict exactly how a
particular AgentSpeak agent would interpret a particular message in a
given situation. Using this as a reference model, it should in
principle be possible to implement communication for other agent
programming languages. Of course, our semantics is not language
independent: it was developed specifically for AgentSpeak, so language
specifics ought to be considered. However, attempts at giving
semantics of agent communication that are language independent have
their own problems, most notably the computational grounding problem
referred to above. Our semantics, while developed specifically for a
practical agent programming language, have the advantage of not
relying on mechanisms (such as abstractly defined mental states) that
cannot be checked for real programs. We note that, to the best of our
knowledge, our work represents the first semantics given for a
speech-act style, ``knowledge level'' communication language that is
used in a real system.

Our current work does not consider commissive and declarative speech
acts. These are surely relevant topics for future work, since
commissive acts and declarations are relevant for various forms of
agent interaction, such as negotiation.  Nevertheless, in the proposed
framework it is possible for the programmer or multi-agent system
designer to incorporate such more elaborate forms of interactions by
writing appropriate plans.

In this work, we assume that communication occurs among agents written
in the same programming language, and cannot be adopted directly in
heterogeneous multi-agent systems. (Consider, for example, the issues
arising in processing an $\ASKHOW$ performative, which involves
sending a plan to another agent.) However, for a variety of other
agent languages, it should not be difficult to write ``wrappers'' for
translating message contents.

Other relevant areas for future investigation are those regarding role
definitions and social structures or agent organisations. We consider
that these would be interesting developments of the proposed
$\SOCACC()$ function and libraries of plans or plan patterns. Deontic
relationships and social norms are also closely related to such
extensions. In the case of e-business, for instance, a contract
usually creates a number of obligations for the contractors.

Future work should also consider giving a better formal treatment of
information sources, in particular for the case of plans being
exchanged between agents. Further communication aspects such as
ontological agreement among \as\ agents, and reasoning about
information sources (e.g., in executing test goals or choosing plans
based on annotations) will also be considered in future work. We
further expect sophisticated multi-agent system applications to be
developed with \as\ interpreters implemented according to our
semantics.


\section*{Acknowledgements}

Many thanks to Jomi F. H\"{u}bner for his comments and suggestions on
an earlier version of this paper, and to Berndt Farwer and Louise
Dennis who carefully proofread it. The first and second authors
acknowledge the support of CNPq.


\begin{thebibliography}{}

\bibitem[\protect\BCAY{Alechina, Bordini, H{\"u}bner, Jago, \BBA\
  Logan}{Alechina et~al.}{2006}]{AlechinaABRAS}
Alechina, N., Bordini, R.~H., H{\"u}bner, J.~F., Jago, M., \BBA\ Logan, B.
  \BBOP2006\BBCP.
\newblock \BBOQ Automating belief revision for agentspeak\BBCQ\
\newblock In Baldoni, M.\BBACOMMA\  \BBA\ Endriss, U.\BEDS, {\Bem Proceedings
  of the Fourth International Workshop on Declarative Agent Languages and
  Technologies (DALT 2006), held with AAMAS 2006, 8th May, Hakodate, Japan},
  \BPGS\ 1--16.

\bibitem[\protect\BCAY{Allen, Hendler, \BBA\ Tate}{Allen
  et~al.}{1990}]{Allen:Readings}
Allen, J.~F., Hendler, J., \BBA\ Tate, A.\BEDS. \BBOP1990\BBCP.
\newblock {\Bem Readings in Planning}.
\newblock Morgan Kaufmann.

\bibitem[\protect\BCAY{Ancona, Mascardi, H{\"u}bner, \BBA\ Bordini}{Ancona
  et~al.}{2004}]{AnconaCooAScape}
Ancona, D., Mascardi, V., H{\"u}bner, J.~F., \BBA\ Bordini, R.~H.
  \BBOP2004\BBCP.
\newblock \BBOQ {Coo-AgentSpeak}: Cooperation in {AgentSpeak} through plan
  exchange\BBCQ\
\newblock In Jennings, N.~R., Sierra, C., Sonenberg, L., \BBA\ Tambe, M.\BEDS,
  {\Bem Proceedings of the Third International Joint Conference on Autonomous
  Agents and Multi-Agent Systems (AAMAS-2004), New York, NY, 19--23 July},
  \BPGS\ 698--705\ New York, NY. ACM Press.

\bibitem[\protect\BCAY{Austin}{Austin}{1962}]{AustinHDTW}
Austin, J.~L. \BBOP1962\BBCP.
\newblock {\Bem How to Do Things with Words}.
\newblock Oxford University Press, London.

\bibitem[\protect\BCAY{Ballmer \BBA\ Brennenstuhl}{Ballmer \BBA\
  Brennenstuhl}{1981}]{BallmerSAC}
Ballmer, T.~T.\BBACOMMA\  \BBA\ Brennenstuhl, W. \BBOP1981\BBCP.
\newblock {\Bem Speech Act Classification: A Study in the Lexical Analysis of
  English Speech Activity Verbs}.
\newblock Springer-Verlag, Berlin.

\bibitem[\protect\BCAY{Bordini, Bazzan, Jannone, Basso, Vicari, \BBA\
  Lesser}{Bordini et~al.}{2002}]{BordiniASXLeisbadtts}
Bordini, R.~H., Bazzan, A. L.~C., Jannone, R.~O., Basso, D.~M., Vicari, R.~M.,
  \BBA\ Lesser, V.~R. \BBOP2002\BBCP.
\newblock \BBOQ {AgentSpeak(XL)}: Efficient intention selection in {BDI} agents
  via decision-theoretic task scheduling\BBCQ\
\newblock In Castelfranchi, C.\BBACOMMA\  \BBA\ Johnson, W.~L.\BEDS, {\Bem
  Proceedings of the First International Joint Conference on Autonomous Agents
  and Multi-Agent Systems (AAMAS-2002), 15--19 July, Bologna, Italy}, \BPGS\
  1294--1302\ New York, NY. ACM Press.

\bibitem[\protect\BCAY{Bordini, da~Rocha~Costa, H{\"u}bner, Moreira, Okuyama,
  \BBA\ Vieira}{Bordini et~al.}{2005}]{BordiniMAS-SOCsspaop}
Bordini, R.~H., da~Rocha~Costa, A.~C., H{\"u}bner, J.~F., Moreira, {\'A}.~F.,
  Okuyama, F.~Y., \BBA\ Vieira, R. \BBOP2005\BBCP.
\newblock \BBOQ {MAS-SOC}: a social simulation platform based on agent-oriented
  programming\BBCQ\
\newblock {\Bem Journal of Artificial Societies and Social Simulation}, {\Bem
  8\/}(3).
\newblock JASSS Forum, {\textless
  http://jasss.soc.surrey.ac.uk/8/3/7.html\textgreater}.

\bibitem[\protect\BCAY{Bordini, Fisher, Pardavila, \BBA\ Wooldridge}{Bordini
  et~al.}{2003}]{BordiniMCAS}
Bordini, R.~H., Fisher, M., Pardavila, C., \BBA\ Wooldridge, M. \BBOP2003\BBCP.
\newblock \BBOQ Model checking {AgentSpeak}\BBCQ\
\newblock In Rosenschein, J.~S., Sandholm, T., Wooldridge, M., \BBA\ Yokoo,
  M.\BEDS, {\Bem Proceedings of the Second International Joint Conference on
  Autonomous Agents and Multi-Agent Systems (AAMAS-2003), Melbourne, Australia,
  14--18 July}, \BPGS\ 409--416\ New York, NY. ACM Press.

\bibitem[\protect\BCAY{Bordini, Fisher, Visser, \BBA\ Wooldridge}{Bordini
  et~al.}{2004}]{BordiniMCRA}
Bordini, R.~H., Fisher, M., Visser, W., \BBA\ Wooldridge, M. \BBOP2004\BBCP.
\newblock \BBOQ Model checking rational agents\BBCQ\
\newblock {\Bem IEEE Intelligent Systems}, {\Bem 19\/}(5), 46--52.

\bibitem[\protect\BCAY{Bordini \BBA\ H{\"u}bner}{Bordini \BBA\
  H{\"u}bner}{2007}]{BordiniJASON}
Bordini, R.~H.\BBACOMMA\  \BBA\ H{\"u}bner, J.~F. \BBOP2007\BBCP.
\newblock {\Bem {\textbf{\textit{{Jason}}}}: A {Java}-based Interpreter for an
  Extended version of {AgentSpeak}\/} (Manual, version 0.9 \BEd).
\newblock {\mbox{\texttt{http://jason.sourceforge.net/}}}.

\bibitem[\protect\BCAY{Bordini, H{\"u}bner, \BBA\ Tralamazza}{Bordini
  et~al.}{2006}]{BordiniUJITGMpd}
Bordini, R.~H., H{\"u}bner, J.~F., \BBA\ Tralamazza, D.~M. \BBOP2006\BBCP.
\newblock \BBOQ Using {{\textbf{\textit{Jason}}}} to implement a team of gold
  miners (a preliminary design)\BBCQ\
\newblock In Inoue, K., Satoh, K., \BBA\ Toni, F.\BEDS, {\Bem Proceedings of
  the Seventh Workshop on Computational Logic in Multi-Agent Systems (CLIMA
  VII), held with AAMAS 2006, 8--9th May, Hakodate, Japan}, \BPGS\ 233--237.
\newblock (Clima Contest paper).

\bibitem[\protect\BCAY{Bordini, H{\"u}bner, \BBA\ Vieira}{Bordini
  et~al.}{2005}]{BordiniJGFAOP}
Bordini, R.~H., H{\"u}bner, J.~F., \BBA\ Vieira, R. \BBOP2005\BBCP.
\newblock \BBOQ {\textbf{\textit{Jason}}} and the {G}olden {F}leece of
  agent-oriented programming\BBCQ\
\newblock In Bordini, R.~H., Dastani, M., Dix, J., \BBA\ El~Fallah~Seghrouchni,
  A.\BEDS, {\Bem Multi-Agent Programming: Languages, Platforms, and
  Applications}, \BCH~1. Springer-Verlag.

\bibitem[\protect\BCAY{Bordini \BBA\ Moreira}{Bordini \BBA\
  Moreira}{2004}]{BordiniPBPAOPL-ATPA}
Bordini, R.~H.\BBACOMMA\  \BBA\ Moreira, {\'A}.~F. \BBOP2004\BBCP.
\newblock \BBOQ Proving {BDI} properties of agent-oriented programming
  languages: The asymmetry thesis principles in {AgentSpeak(L)}\BBCQ\
\newblock {\Bem Annals of Mathematics and Artificial Intelligence}, {\Bem
  42\/}(1--3), 197--226.
\newblock Special Issue on Computational Logic in Multi-Agent Systems.

\bibitem[\protect\BCAY{Bordini, Visser, Fisher, Pardavila, \BBA\
  Wooldridge}{Bordini et~al.}{2003}]{BordiniMCMAP}
Bordini, R.~H., Visser, W., Fisher, M., Pardavila, C., \BBA\ Wooldridge, M.
  \BBOP2003\BBCP.
\newblock \BBOQ Model checking multi-agent programs with {CASP}\BBCQ\
\newblock In Hunt~Jr., W.~A.\BBACOMMA\  \BBA\ Somenzi, F.\BEDS, {\Bem
  Proceedgins of the Fifteenth Conference on Computer-Aided Verification
  (CAV-2003), Boulder, CO, 8--12 July}, \lowercase{\BNUM}\ 2725 in LNCS, \BPGS\
  110--113\ Berlin. Springer-Verlag.
\newblock Tool description.

\bibitem[\protect\BCAY{Bratman}{Bratman}{1987}]{BratmanIPPR}
Bratman, M.~E. \BBOP1987\BBCP.
\newblock {\Bem Intentions, Plans and Practical Reason}.
\newblock Harvard University Press, Cambridge, MA.

\bibitem[\protect\BCAY{Castelfranchi \BBA\ Falcone}{Castelfranchi \BBA\
  Falcone}{1998}]{CastPTMASCASIQ}
Castelfranchi, C.\BBACOMMA\  \BBA\ Falcone, R. \BBOP1998\BBCP.
\newblock \BBOQ Principles of trust for {MAS}: Cognitive anatomy, social
  importance, and quantification\BBCQ\
\newblock In Demazeau, Y.\BED, {\Bem Proceedings of the Third International
  Conference on Multi-Agent Systems (ICMAS'98), 4--7 July, Paris}, \BPGS\
  72--79\ Washington. IEEE Computer Society Press.

\bibitem[\protect\BCAY{Cohen \BBA\ Perrault}{Cohen \BBA\
  Perrault}{1979}]{cohen:79a}
Cohen, P.\BBACOMMA\  \BBA\ Perrault, R. \BBOP1979\BBCP.
\newblock \BBOQ Elements of a plan based theory of speech acts\BBCQ\
\newblock {\Bem Cognitive Science}, {\Bem 3}, 177--212.

\bibitem[\protect\BCAY{Cohen \BBA\ Levesque}{Cohen \BBA\
  Levesque}{1990a}]{CohenICC}
Cohen, P.~R.\BBACOMMA\  \BBA\ Levesque, H.~J. \BBOP1990a\BBCP.
\newblock \BBOQ Intention is choice with commitment\BBCQ\
\newblock {\Bem Artificial Intelligence}, {\Bem 42\/}(3), 213--261.

\bibitem[\protect\BCAY{Cohen \BBA\ Levesque}{Cohen \BBA\
  Levesque}{1990b}]{CohenRIBC}
Cohen, P.~R.\BBACOMMA\  \BBA\ Levesque, H.~J. \BBOP1990b\BBCP.
\newblock \BBOQ Rational interaction as the basis for communication\BBCQ\
\newblock In Cohen, P.~R., Morgan, J., \BBA\ Pollack, M.~E.\BEDS, {\Bem
  Intentions in Communication}, \BCH~12, \BPGS\ 221--255. MIT Press, Cambridge,
  MA.

\bibitem[\protect\BCAY{Dastani, van~der Ham, \BBA\ Dignum}{Dastani
  et~al.}{2003}]{DastaniCGDA}
Dastani, M., van~der Ham, J., \BBA\ Dignum, F. \BBOP2003\BBCP.
\newblock \BBOQ Communication for goal directed agents\BBCQ\
\newblock In Huget, M.-P.\BED, {\Bem Communication in Multiagent Systems},
  \lowercase{\BVOL}\ 2650 of {\Bem LNCS}, \BPGS\ 239--252. Springer-Verlag.

\bibitem[\protect\BCAY{de~Boer, van Eijk, Van Der~Hoek, \BBA\ Meyer}{de~Boer
  et~al.}{2000}]{BoerFSEIMAS}
de~Boer, F.~S., van Eijk, R.~M., Van Der~Hoek, W., \BBA\ Meyer, J.-J.~C.
  \BBOP2000\BBCP.
\newblock \BBOQ Failure semantics for the exchange of information in
  multi-agent systems\BBCQ\
\newblock In Palamidessi, C.\BED, {\Bem Eleventh International Conference on
  Concurrency Theory (CONCUR 2000), University Park, PA, 22--25 August},
  \lowercase{\BNUM}\ 1877 in LNCS, \BPGS\ 214--228. Springer-Verlag.

\bibitem[\protect\BCAY{Doran \BBA\ Gilbert}{Doran \BBA\
  Gilbert}{1994}]{DoranSSI}
Doran, J.\BBACOMMA\  \BBA\ Gilbert, N. \BBOP1994\BBCP.
\newblock \BBOQ Simulating societies: An introduction\BBCQ\
\newblock In Gilbert, N.\BBACOMMA\  \BBA\ Doran, J.\BEDS, {\Bem Simulating
  Society: The Computer Simulation ofSocial Phenomena}, \BCH~1, \BPGS\ 1--18.
  UCL Press, London.

\bibitem[\protect\BCAY{Genesereth \BBA\ Ketchpel}{Genesereth \BBA\
  Ketchpel}{1994}]{GeneSA}
Genesereth, M.~R.\BBACOMMA\  \BBA\ Ketchpel, S.~P. \BBOP1994\BBCP.
\newblock \BBOQ Software agents\BBCQ\
\newblock {\Bem Communications of the ACM}, {\Bem 37\/}(7), 48--53.

\bibitem[\protect\BCAY{Georgeff \BBA\ Lansky}{Georgeff \BBA\
  Lansky}{1987}]{GeorRRP}
Georgeff, M.~P.\BBACOMMA\  \BBA\ Lansky, A.~L. \BBOP1987\BBCP.
\newblock \BBOQ Reactive reasoning and planning\BBCQ\
\newblock In {\Bem Proceedings of the Sixth National Conference on Artificial
  Intelligence (AAAI'87), 13--17 July,1987, Seattle, WA}, \BPGS\ 677--682\
  Manlo Park, CA. AAAI Press~/ MIT Press.

\bibitem[\protect\BCAY{Ghallab, Nau, \BBA\ Traverso}{Ghallab
  et~al.}{2004}]{ghallab:2004a}
Ghallab, M., Nau, D., \BBA\ Traverso, P. \BBOP2004\BBCP.
\newblock {\Bem Automated Planning: Theory and Practice}.
\newblock Morgan Kaufmann.

\bibitem[\protect\BCAY{Guerin \BBA\ Pitt}{Guerin \BBA\ Pitt}{2001}]{Guerin01}
Guerin, F.\BBACOMMA\  \BBA\ Pitt, J. \BBOP2001\BBCP.
\newblock \BBOQ Denotational semantics for agent communication language\BBCQ\
\newblock In {\Bem Proceedings of the fifth international conference on
  Autonomous Agents (Agents 2001), 28th May -- 1st June, Montreal Canada},
  \BPGS\ 497--504. ACM Press.

\bibitem[\protect\BCAY{Halpern}{Halpern}{1990}]{halpern:90a}
Halpern, J.~Y. \BBOP1990\BBCP.
\newblock \BBOQ Knowledge and common knowledge in a distributed
  environment\BBCQ\
\newblock {\Bem Journal of the ACM}, {\Bem 37\/}(3).

\bibitem[\protect\BCAY{Halpern \BBA\ Zuck}{Halpern \BBA\
  Zuck}{1992}]{halpern:92b}
Halpern, J.~Y.\BBACOMMA\  \BBA\ Zuck, L.~D. \BBOP1992\BBCP.
\newblock \BBOQ A little knowledge goes a long way: knowledge-based derivations
  and correctness proofs for a family of protocols\BBCQ\
\newblock {\Bem Journal of the ACM}, {\Bem 39\/}(3), 449--478.

\bibitem[\protect\BCAY{H{\"u}bner, Sichman, \BBA\ Boissier}{H{\"u}bner
  et~al.}{2004}]{HubnerUMCFMR}
H{\"u}bner, J.~F., Sichman, J.~S., \BBA\ Boissier, O. \BBOP2004\BBCP.
\newblock \BBOQ Using the {$\mathcal{M}$oise+} for a cooperative framework of
  {MAS} reorganisation.\BBCQ\
\newblock In Bazzan, A. L.~C.\BBACOMMA\  \BBA\ Labidi, S.\BEDS, {\Bem Advances
  in Artificial Intelligence - SBIA 2004, 17th Brazilian Symposium on
  Artificial Intelligence, S{\~a}o Luis, Maranh{\~a}o, Brazil, September 29 -
  October 1, 2004, Proceedings}, \lowercase{\BVOL}\ 3171 of {\Bem LNCS}, \BPGS\
  506--515. Springer-Verlag.

\bibitem[\protect\BCAY{Jennings}{Jennings}{1995}]{jennings:95a}
Jennings, N.~R. \BBOP1995\BBCP.
\newblock \BBOQ Controlling cooperative problem solving in industrial
  multi-agent systems using joint intentions\BBCQ\
\newblock {\Bem Artificial Intelligence}, {\Bem 75\/}(2), 195--240.

\bibitem[\protect\BCAY{Kinny}{Kinny}{1993}]{KinnyDMARSALS}
Kinny, D. \BBOP1993\BBCP.
\newblock \BBOQ The distributed multi-agent reasoning system architecture and
  language specification\BBCQ\
\newblock \BTR, Australian Artificial Intelligence Institute, Melbourne,
  Australia.

\bibitem[\protect\BCAY{Krafta, de~Oliveira, \BBA\ Bordini}{Krafta
  et~al.}{2003}]{KraftaCOHA}
Krafta, R., de~Oliveira, D., \BBA\ Bordini, R.~H. \BBOP2003\BBCP.
\newblock \BBOQ The city as object of human agency\BBCQ\
\newblock In {\Bem Fourth International Space Syntax Symposium (SSS4), London,
  17--19 June}, \BPGS\ 33.1--33.18.

\bibitem[\protect\BCAY{Kumar \BBA\ Cohen}{Kumar \BBA\
  Cohen}{2000}]{KumarCohen00}
Kumar, S.\BBACOMMA\  \BBA\ Cohen, P.~R. \BBOP2000\BBCP.
\newblock \BBOQ Towards a fault-tolerant multi-agent system architecture\BBCQ\
\newblock In {\Bem Proceedings of the Fourth International Conference on
  Autonomous Agents (Agents 2000), 3--7 June, Barcelona, Spain}, \BPGS\
  459--466. ACM Press.

\bibitem[\protect\BCAY{Labrou \BBA\ Finin}{Labrou \BBA\
  Finin}{1994}]{LabrouSAKQML}
Labrou, Y.\BBACOMMA\  \BBA\ Finin, T. \BBOP1994\BBCP.
\newblock \BBOQ A semantics approach for {KQML}---a general purpose
  communication language for software agents\BBCQ\
\newblock In {\Bem Proceedings of the Third International Conference on
  Information and Knowledge Management ({CIKM'94}), 29th November -- 2nd
  December, Gaithersburg, MD}. ACM Press.

\bibitem[\protect\BCAY{Labrou, Finin, \BBA\ Peng}{Labrou
  et~al.}{1999}]{LabrouCLACL}
Labrou, Y., Finin, T., \BBA\ Peng, Y. \BBOP1999\BBCP.
\newblock \BBOQ The current landscape of agent communication languages\BBCQ\
\newblock {\Bem Intelligent Systems}, {\Bem 14\/}(2), 45--52.

\bibitem[\protect\BCAY{Levesque, Cohen, \BBA\ Nunes}{Levesque
  et~al.}{1990}]{levesque:90a}
Levesque, H.~J., Cohen, P.~R., \BBA\ Nunes, J. H.~T. \BBOP1990\BBCP.
\newblock \BBOQ On acting together\BBCQ\
\newblock In {\Bem Proceedings of the Eighth National Conference on Artificial
  Intelligence (AAAI-1990), 29th July -- 3rd August, Boston, MA}, \BPGS\
  94--99. AAAI Press.

\bibitem[\protect\BCAY{Levinson}{Levinson}{1981}]{LeviEISAMD}
Levinson, S.~C. \BBOP1981\BBCP.
\newblock \BBOQ The essential inadequacies of speech act models of
  dialogue\BBCQ\
\newblock In Parret, H., Sbisa, M., \BBA\ Verschuren, J.\BEDS, {\Bem
  Possibilities and limitations of pragmatics: Proceedings of the Conference on
  Pragmatics at Urbino, July, 1979}, \BPGS\ 473--492. Benjamins, Amsterdam.

\bibitem[\protect\BCAY{Mayfield, Labrou, \BBA\ Finin}{Mayfield
  et~al.}{1996}]{MayfieldEKQMLACL}
Mayfield, J., Labrou, Y., \BBA\ Finin, T. \BBOP1996\BBCP.
\newblock \BBOQ Evaluation of {KQML} as an agent communication language\BBCQ\
\newblock In Wooldridge, M., M{\"u}ller, J.~P., \BBA\ Tambe, M.\BEDS, {\Bem
  Intelligent Agents~II---Proceedings of the Second International Workshop on
  Agent Theories, Architectures, and Languages (ATAL'95), held as part of
  IJCAI'95, Montr{\'e}al, Canada, August1995}, \lowercase{\BNUM}\ 1037 in LNAI,
  \BPGS\ 347--360\ Berlin. Springer-Verlag.

\bibitem[\protect\BCAY{Moreira \BBA\ Bordini}{Moreira \BBA\
  Bordini}{2002}]{MoreiraOSBAOPL}
Moreira, {\'A}.~F.\BBACOMMA\  \BBA\ Bordini, R.~H. \BBOP2002\BBCP.
\newblock \BBOQ An operational semantics for a {BDI} agent-oriented programming
  language\BBCQ\
\newblock In Meyer, J.-J.~C.\BBACOMMA\  \BBA\ Wooldridge, M.~J.\BEDS, {\Bem
  Proceedings of the Workshop on Logics for Agent-Based Systems (LABS-02), held
  in conjunction with the Eighth International Conference on Principles of
  Knowledge Representation and Reasoning (KR2002), April 22--25, Toulouse,
  France}, \BPGS\ 45--59.

\bibitem[\protect\BCAY{Moreira, Vieira, \BBA\ Bordini}{Moreira
  et~al.}{2004}]{MoreiraEOSBAOPLISABC-LNAI}
Moreira, {\'A}.~F., Vieira, R., \BBA\ Bordini, R.~H. \BBOP2004\BBCP.
\newblock \BBOQ Extending the operational semantics of a {BDI} agent-oriented
  programming language for introducing speech-act based communication\BBCQ\
\newblock In Leite, J., Omicini, A., Sterling, L., \BBA\ Torroni, P.\BEDS,
  {\Bem Declarative Agent Languages and Technologies, Proceedings of the First
  International Workshop (DALT-03), held with AAMAS-03, 15 July, 2003,
  Melbourne, Australia (Revised Selected and Invited Papers)},
  \lowercase{\BNUM}\ 2990 in LNAI, \BPGS\ 135--154\ Berlin. Springer-Verlag.

\bibitem[\protect\BCAY{Moreira, Vieira, Bordini, \BBA\ H{\"u}bner}{Moreira
  et~al.}{2006}]{MoreiraAOPUOR-LNCS}
Moreira, {\'A}.~F., Vieira, R., Bordini, R.~H., \BBA\ H{\"u}bner, J.~F.
  \BBOP2006\BBCP.
\newblock \BBOQ Agent-oriented programming with underlying ontological
  reasoning\BBCQ\
\newblock In Baldoni, M., Endriss, U., Omicini, A., \BBA\ Torroni, P.\BEDS,
  {\Bem Proceedings of the Third International Workshop on Declarative Agent
  Languages and Technologies (DALT-05), held with AAMAS-05, 25th of July,
  Utrecht, Netherlands}, \lowercase{\BNUM}\ 3904 in LNCS, \BPGS\ 155--170.
  Springer-Verlag.

\bibitem[\protect\BCAY{Okuyama, Bordini, \BBA\ da~Rocha~Costa}{Okuyama
  et~al.}{2005}]{OkuyamaELMSedlmas-LNAI}
Okuyama, F.~Y., Bordini, R.~H., \BBA\ da~Rocha~Costa, A.~C. \BBOP2005\BBCP.
\newblock \BBOQ {ELMS:} an environment description language for multi-agent
  simulations\BBCQ\
\newblock In Weyns, D., van Dyke~Parunak, H., Michel, F., Holvoet, T., \BBA\
  Ferber, J.\BEDS, {\Bem Environments for Multiagent Systems, State-of-the-art
  and Research Challenges. Proceedings of the First International Workshop on
  Environments for Multiagent Systems (E4MAS), held with AAMAS-04, 19th of
  July}, \lowercase{\BNUM}\ 3374 in LNAI, \BPGS\ 91--108\ Berlin.
  Springer-Verlag.

\bibitem[\protect\BCAY{Plotkin}{Plotkin}{1981}]{Plotkin:SOS}
Plotkin, G. \BBOP1981\BBCP.
\newblock \BBOQ A structural approach to operational semantics\BBCQ.
\newblock Technical Report, Department of Computer Science, Aarhus University.

\bibitem[\protect\BCAY{Rao}{Rao}{1996}]{RaoASLBDIASOLCL}
Rao, A.~S. \BBOP1996\BBCP.
\newblock \BBOQ {AgentSpeak(L)}: {BDI} agents speak out in a logical computable
  language\BBCQ\
\newblock In van~de Velde, W.\BBACOMMA\  \BBA\ Perram, J.\BEDS, {\Bem
  Proceedings of the 7th Workshop on Modelling Autonomous Agents in a
  Multi-Agent World (MAAMAW'96), 22--25 January, Eindhoven, The Netherlands},
  \lowercase{\BNUM}\ 1038 in LNAI, \BPGS\ 42--55\ London. Springer-Verlag.

\bibitem[\protect\BCAY{Rao \BBA\ Georgeff}{Rao \BBA\
  Georgeff}{1998}]{RaoDPBDIL}
Rao, A.~S.\BBACOMMA\  \BBA\ Georgeff, M.~P. \BBOP1998\BBCP.
\newblock \BBOQ Decision procedures for {BDI} logics\BBCQ\
\newblock {\Bem Journal of Logic and Computation}, {\Bem 8\/}(3), 293--343.

\bibitem[\protect\BCAY{Searle}{Searle}{1969}]{SearleSAEPL}
Searle, J.~R. \BBOP1969\BBCP.
\newblock {\Bem Speech Acts: An Essay in the Philosophy of Language}.
\newblock Cambridge University Press, Cambridge.

\bibitem[\protect\BCAY{Singh}{Singh}{1994}]{SinghMSTFIKHC}
Singh, M.~P. \BBOP1994\BBCP.
\newblock {\Bem Multiagent Systems---{A} Theoretic Framework for Intentions,
  Know-How, and Communications}.
\newblock \BNUM\ 799 in LNAI. Springer-Verlag, Berlin.

\bibitem[\protect\BCAY{Singh}{Singh}{1998}]{SinghACLRP}
Singh, M.~P. \BBOP1998\BBCP.
\newblock \BBOQ Agent communication languages: Rethinking the principles\BBCQ\
\newblock {\Bem IEEE Computer}, {\Bem 31\/}(12), 40--47.

\bibitem[\protect\BCAY{Smith}{Smith}{1980}]{SmithCNPHLCCDPS}
Smith, R.~G. \BBOP1980\BBCP.
\newblock \BBOQ The contract net protocol: High-level communication and control
  in a distributed problem solver\BBCQ\
\newblock {\Bem IEEE Transactions on Computers}, {\Bem {\textsc{c}}-29\/}(12),
  1104--1113.

\bibitem[\protect\BCAY{Tambe}{Tambe}{1997}]{tambe:97a}
Tambe, M. \BBOP1997\BBCP.
\newblock \BBOQ Towards flexible teamwork\BBCQ\
\newblock {\Bem Journal of Artificial Intelligence Research}, {\Bem 7},
  83--124.

\bibitem[\protect\BCAY{Torres, Nedel, \BBA\ Bordini}{Torres
  et~al.}{2004}]{TorresAAMFAVE}
Torres, J.~A., Nedel, L.~P., \BBA\ Bordini, R.~H. \BBOP2004\BBCP.
\newblock \BBOQ Autonomous agents with multiple foci of attention in virtual
  environments\BBCQ\
\newblock In {\Bem Proceedings of 17th International Conference on Computer
  Animation and Social Agents (CASA 2004), Geneva, Switzerland, 7--9 July},
  \BPGS\ 189--196.

\bibitem[\protect\BCAY{van Eijk, de~Boer, Van Der~Hoek, \BBA\ Meyer}{van Eijk
  et~al.}{2003}]{EijkVFAC}
van Eijk, R.~M., de~Boer, F.~S., Van Der~Hoek, W., \BBA\ Meyer, J.-J.~C.
  \BBOP2003\BBCP.
\newblock \BBOQ A verification framework for agent communication\BBCQ\
\newblock {\Bem Autonomous Agents and Multi-Agent Systems}, {\Bem 6\/}(2),
  185--219.

\bibitem[\protect\BCAY{Vieira, Moreira, Bordini, \BBA\ H{\"u}bner}{Vieira
  et~al.}{2006}]{VieiraAOPLCC}
Vieira, R., Moreira, {\'A}.~F., Bordini, R.~H., \BBA\ H{\"u}bner, J.
  \BBOP2006\BBCP.
\newblock \BBOQ An agent-oriented programming language for computing in
  context\BBCQ\
\newblock In Debenham, J.\BED, {\Bem Proceedings of Second IFIP Symposium on
  Professional Practice in Artificial Intelligence, held with the 19th IFIP
  World Computer Congress, TC-12 Professional Practice Stream, 21--24 August,
  Santiago, Chile}, \lowercase{\BNUM}\ 218 in IFIP, \BPGS\ 61--70\ Berlin.
  Springer-Verlag.

\bibitem[\protect\BCAY{Wooldridge}{Wooldridge}{1998}]{Wool98}
Wooldridge, M. \BBOP1998\BBCP.
\newblock \BBOQ Verifiable semantics for agent communication languages\BBCQ\
\newblock In {\Bem Proceedings of the Third International Conference on
  Multi-Agent Systems (ICMAS'98), 4--7 July, Paris}, \BPGS\ 349--365. IEEE
  Computer Society Press.

\bibitem[\protect\BCAY{Wooldridge}{Wooldridge}{2000a}]{WoolCGTA}
Wooldridge, M. \BBOP2000a\BBCP.
\newblock \BBOQ Computationally grounded theories of agency\BBCQ\
\newblock In Durfee, E.\BED, {\Bem Proceedings of the Fourth International
  Conference on Multi-Agent Systems (ICMAS-2000),10--12 July, Boston}, \BPGS\
  13--20\ Los Alamitos, CA. IEEE Computer Society.
\newblock Paper for an Invited Talk.

\bibitem[\protect\BCAY{Wooldridge}{Wooldridge}{2000b}]{WoolRRA}
Wooldridge, M. \BBOP2000b\BBCP.
\newblock {\Bem Reasoning about Rational Agents}.
\newblock The MIT Press, Cambridge, MA.

\bibitem[\protect\BCAY{Wooldridge}{Wooldridge}{2000c}]{WoolSIVACL}
Wooldridge, M. \BBOP2000c\BBCP.
\newblock \BBOQ Semantic issues in the verification of agent communication
  languages\BBCQ\
\newblock {\Bem Autonomous Agents and Multi-Agent Systems}, {\Bem 3\/}(1),
  9--31.

\bibitem[\protect\BCAY{Wooldridge}{Wooldridge}{2002}]{WoolIMAS}
Wooldridge, M. \BBOP2002\BBCP.
\newblock {\Bem An Introduction to MultiAgent Systems}.
\newblock John Wiley \& Sons.

\end{thebibliography}

\end{document}